%% file: nondetAC_ver2_post_review.tex
\documentclass[final,12pt]{clear2025} % Include author names

\title[Actual Causation and Nondeterministic Causal Models]{Actual Causation and Nondeterministic Causal Models}
\usepackage{times}
\usepackage{tikz}
\usetikzlibrary{arrows}
\usepackage{multicol}
\input{defn}

\newtheorem*{theorem*}{Theorem}

\newtheorem*{proposition*}{Proposition}
\newtheorem*{example*}{Example}

\clearauthor{%
 \Name{Sander Beckers} \Email{srekcebrednas@gmail.com}\\
 \addr Cornell University
}
\begin{document}

\maketitle

\begin{abstract}%
In \citep{beckers25} I introduced nondeterministic causal models as a generalization of Pearl's standard deterministic causal models. I here take advantage of the increased expressivity offered by these models to offer a novel definition of {\em actual causation} (that also applies to deterministic models). Instead of motivating the definition by way of (often subjective) intuitions about examples, I proceed by developing it based entirely on the unique function that it can fulfil in communicating and learning a causal model. First I generalize the more basic notion of {\em counterfactual dependence}, second I show how this notion has a vital role to play in the logic of causal discovery, third I introduce the notion of a {\em structural simplification} of a causal model, and lastly I bring both notions together in my definition of actual causation. Although novel, the resulting definition arrives at verdicts that are almost identical to those of my previous definition \citep{beckers21c,beckers22a}.
%The resulting nondeterministic semantics for causal counterfactuals allows for distinguishing between so-called {\em would}-counterfactuals and {\em might}-counterfactuals. In this paper we use the latter to define nondeterministic counterfactual dependence, and argue that it offers an improved foundation for an interventionist account of {\em actual causation}, which is the relation that holds when a particular event causes another. Concretely, the increased expressivity of nondeterministic models allows us to define that one causal setting is a {\em conservative reduction} of another, meaning that the former increases the nondeterministic possibilities present in the latter by reducing the number of edges in the causal graph in a manner that remains consistent with the actual observations. We define actual causation as counterfactual dependence in some conservative reduction, and show how this definition forms an improvement over earlier work. Importantly, the definition applies just as well to standard, deterministic, models.
\end{abstract}

\begin{keywords}%
  nondeterminism; counterfactuals; actual causation; counterfactual dependence%
\end{keywords}

\section{Introduction}

In \citep{beckers25} I introduced {\em Nondeterministic Structural Causal Models} (NSCMs) as a generalization of \cite{pearl:book2}'s influential Deterministic Structural Causal Models (DSCMs). DSCMs form the mathematical bedrock of Pearl's widely influential causal modeling framework. These causal models are deterministic in the sense that they assume the values of causal parents uniquely determine the values of their children, both actually and counterfactually. As I argue in \citep{beckers25}, both assumptions are unnecessarily restrictive, and nondeterministic causal models are the result of dropping them. 

I here take advantage of the increased expressivity offered by NSCMs to offer a novel definition of {\em actual causation} (that applies to both NSCMs and DSCMs), which is the causal relation that obtains between specific instances of events, such as when a specific patient's taking aspirin causes their headache to disappear. Not only has this concept been notoriously hard to define (in comparison to other causal relations, such as the average causal effect, for example), it has also proven difficult to objectively evaluate the quality of different proposed definitions \citep{stonesoup, halpernbook, beckers21c}. Instead of starting from paradigmatic test cases that a definition should ``get right'', I here develop a definition by focussing exclusively on the vital function that statements of actual causation fulfil in efficiently communicating and learning a causal model. 

The paper proceeds as follows. The next section formally introduces NSCMs (Sec. \ref{sec:cm}), followed by an informal summary and motivation for the main idea behind my definition of actual causation (Sec. \ref{sec:inf}). This sets the stage for taking all the necessary steps towards formally developing this idea, starting with an investigation of the more basic concept of {\em counterfactual dependence} in the context of NSCMs (Sec. \ref{sec:cd}). This reveals a hitherto ignored notion of {\em interventional dependence} (Sec. \ref{sec:id}) that can play a crucial role in the causal discovery of causal models from an idealized logical perspective (Sec. \ref{sec:disc}). I then  introduce the idea of structurally simplifying a causal model (Sec. \ref{sec:ss}), to arrive at a definition of actual causation as the presence of counterfactual dependence in some {\em structural simplification} of a causal model (Sec. \ref{sec:ac}).
\iffalse
To this end, I start (Sec. \ref{sec:cd}) with investigating the more basic concept of {\em counterfactual dependence} in the context of NSCMs. This reveals a hitherto ignored notion of {\em interventional dependence} (Sec. \ref{sec:id}) that can play a crucial role in the causal discovery of causal models from an idealized logical perspective (Sec. \ref{sec:disc}). I then introduce the idea of structurally simplifying a causal model (Sec. \ref{sec:ss}), to arrive at a definition of actual causation as the presence of counterfactual dependence in some {\em structural simplification} of a causal model (Sec. \ref{sec:ac}). To start, I formally introduce NSCMs.
\fi

\section{Nondeterministic Structural Causal Models\protect\footnote{This section is based almost ad verbatim on \citep{beckers25}.}}\label{sec:cm}
 %We first define a signature as the variables out of which a causal model is built up. 
\begin{definition}
A signature $\cal S$ is a tuple $(\U,\V,\R)$, where $\U$
is a set of \emph{exogenous} variables, $\V$ is a set 
of \emph{endogenous} variables,
and $\R$ a function that associates with every variable $Y \in  
\U \union \V$ a nonempty set $\R(Y)$ of possible values for $Y$
(i.e., the set of values over which $Y$ {\em ranges}).
If $\vec{X} = (X_1, \ldots, X_n)$, $\R(\vec{X})$ denotes the
Cartesian product $\R(X_1) \times \cdots \times \R(X_n)$. \end{definition}
Nondeterministic models generalize \cite{halpern00,halpernbook}'s definition of deterministic causal models by using multi-valued functions. In addition to using multi-valued functions, we depart from Halpern by explicitly including the causal graph as an element of the causal model. 
\begin{definition}
A \emph{causal model} (or a {\em Nondeterministic Structural Causal Model -- NSCM}) $M$ is a triple $(\cal S,\F, \G)$, 
where $\cal S$ is a signature, $\G$ is a directed acyclic graph -- a DAG -- such that there is one node for each variable in $\cal S$, and
$\F$ associates with each endogenous
variable $X$ a \emph{structural equation} $F_X$ that takes on the form $X= f_X(\vec{Pa_X})$, where $\vec{Pa_X} \subseteq (\U \cup \V)$ are the {\em parents} of $X$ in $\G$, and $f_X: \R(\vec{Pa_X}) \rightarrow \P(\R(X))_{\emptyset}$. (Here $\P(\R(X))_{\emptyset}$ is the {\em powerset} of $\R(X)$: the set that contains as its elements all subsets of $\R(X)$, except for $\emptyset$.)\end{definition}
\iffalse If for each $f_X$ the co-domain does not contain the empty set, we say that a causal model is {\em total}. We here restrict ourselves to total causal models, so that for all possible settings $\vec{pa_X}$ of the parents, there exists at least one solution $x$ for the child. Furthermore,\fi
It will prove useful to adopt Halpern's -- by now standard -- graphical convention that for any {\em deterministic} equation $Y=f_Y(\vec{Pa_Y})$, $Y$ functionally depends on each of its parents $X \in \vec{Pa_Y}$, where functional dependence means that there exists some setting $\vec{z}$ of all the other parents and values $x,x'$ such that $f_Y(\vec{z},x) \neq f_Y(\vec{z},x')$. %TODO: (This can be justified by simplicity: both models will satisfy identical equations, but the one is simpler.)

The exogenous variables $\U$ are taken to represent the background conditions that are simply given. We call $\vec{u} \in \R(\U)$ a {\em context}, $\vec{v} \in \R(\V)$ a {\em state}, a $(\vec{u},\vec{v}) \in \R(\U \cup \V)$ is a {\em world}, and $(M,\vec{u},\vec{v})$ a {\em causal setting}. In {\em deterministic} causal models all the functions $f_X$ are standard as opposed to multi-valued, and thus each equation has a unique solution $x$ for each choice of values $\vec{pa_X}$. In nondeterministic models, a {\em solution} of the equation $X=f_X(\vec{Pa_X})$ is a tuple $(x,\vec{pa_X})$ such that $x \in f_X(\vec{pa_X})$. A solution of $M$ is a world $(\vec{u},\vec{v})$ that is a solution of all equations in $\F$.

\subsection{The Causal Language and Semantics}\label{sec:lan}

Given a signature $\cal S = (\U,\V,\R)$, an \emph{atomic formula} is a
formula of the form $X = x$, for  $X \in \V$ and $x \in \R(X)$.  
A {\em basic formula (over $\cal S$)\/} $\phi$ is a Boolean combination of atomic formulas. An \emph{intervention} has the form $\vec{Y} \gets \vec{y}$, where $Y_1, \ldots, Y_k$ are distinct variables in $\V$, and $y_i \in \R(Y_i)$ for each $1 \leq i \leq k$.
 A {\em basic causal formula} has the form  $[Y_1 \gets y_1, \ldots, Y_k \gets y_k] \phi$, where $\phi$ is a basic formula and $Y_1,\ldots,Y_k$ are distinct variables in $\V$. Such a formula is abbreviated as $[\vec{Y} \gets \vec{y}]\phi$. The special case where $k=0$ is abbreviated as $\phi$.\footnote{This is warranted because $\phi$ and $[]\phi$ turn out to be equivalent, just as was the case for \cite{halpern00,halpernbook}.} Finally, a {\em causal formula} is a Boolean combination of basic causal formulas. The language $\cal{L}(\cal{S})$ that we consider consists of all causal formulas.
 
A causal formula $\psi$ is true or false in a causal setting. We write $(M,\vec{u},\vec{v}) \sat \psi$  if the causal formula $\psi$ is true in causal model $M$ given world $(\vec{u},\vec{v})$. %We call a model-world pair $(M,\vec{u},\vec{v})$ a {\em causal setting}.

We first define the $\sat$ relation for basic formulas. $(M,\vec{u},\vec{v}) \sat X=x$ if 
%$(\vec{u},\vec{v})$ is a solution of $M$ and 
$x$ is the restriction of $\vec{v}$ to $X$. We extend $\sat$ to basic formulas $\phi$ in the standard way. Note that the truth of basic formulas is determined solely by the state $\vec{v}$, and thus we often also write $\vec{v} \sat \phi$. 

In order to define the $\sat$ relation for causal formulas, we introduce two operations on a causal model, the {\em actualized refinement} that is the result of integrating the actual behavior of the equations as observed in a world $(\vec{u},\vec{v})$ into the equations of a model $M$, and the {\em intervened} model that is the result of performing an intervention on the equations of a model $M$. Given $\vec{X}$ and $\vec{Y}$, we let $\vec{X}_{\vec{Y}}$ denote the restriction of $\vec{X}$ to $\vec{X} \cap \vec{Y}$, and thus $\vec{x}_{\vec{Y}} \in \R(\vec{X} \cap \vec{Y})$.

\begin{definition}\label{def:ar} Given a solution $(\vec{u},\vec{v})$ of a model $M = (\cal S,\F,\G)$, we define the {\em actualized refinement} $M^{(\vec{u},\vec{v})}$ as the model in which $\F$ is
replaced by $\F^{(\vec{u},\vec{v})}$, as follows: for each variable $X \in \V$, its function $f_X$  is replaced by $f^{(\vec{pa_X},x)}_X$ that behaves identically to $f_X$ for all inputs except for $\vec{pa_X}$. Instead, $f^{(\vec{pa_X},x)}_X(\vec{pa_X})=x$. Here $x=(\vec{u},\vec{v})_X$ and $\vec{pa_X}=(\vec{u},\vec{v})_{\vec{Pa_X}}$. 
\end{definition}

Setting the value of some variables $\vec{Y}$ to $\vec{y}$ in a causal 
model $M = (\cal S,\F,\G)$ results in a new causal model, denoted $M_{\vec{Y}
\gets \vec{y}}$, which is identical to $M$, except that $\F$ is
replaced by $\F^{\vec{Y} \gets \vec{y}}$: for each variable $X \notin
  \vec{Y}$, $F^{\vec{Y} \gets \vec{y}}_X = F_X$ (i.e., the equation
  for $X$ is unchanged), while for
each $Y'$ in $\vec{Y}$, the equation $F_{Y'}$ for $Y'$ is replaced by $Y' =\vec{y}_{Y'}$. Similarly, $\G$ is replaced with $\G^{\vec{Y} \gets \vec{y}}$.

With these operations in place, we can define the $\sat$ relation for basic causal formulas, relative to settings  $(M,\vec{u},\vec{v})$ such that $(\vec{u},\vec{v})$ is a solution of $M$. $(M,\vec{u},\vec{v}) \sat [\vec{Y} \gets \vec{y}]\phi$ iff $\vec{v}' \sat \phi$ for all states $\vec{v}'$ such that $(\vec{u},\vec{v}')$ is a solution of $(M^{(\vec{u},\vec{v})})_{\vec{Y} \gets \vec{y}}$. We inductively extend the semantics to causal formulas in the standard way, that is,  $(M,\vec{u},\vec{v}) \sat [\vec{Y} \gets \vec{y}]\phi_1 \land [\vec{Z} \gets \vec{z}]\phi_2$ iff $(M,\vec{u},\vec{v}) \sat [\vec{Y} \gets \vec{y}]\phi_1$ and $(M,\vec{u},\vec{v}) \sat [\vec{Z} \gets \vec{z}]\phi_2$, and similarly for $\lnot $ and $\lor$.

We define $\langle \vec{Y} \gets \vec{y} \rangle \phi$ as an abbreviation of  $\neg [\vec{Y} \gets \vec{y}] \neg \phi$. So $(M,\vec{u},\vec{v}) \sat \langle \vec{Y} \gets \vec{y} \rangle \phi$ iff $\vec{v}' \sat \phi$ for some solution $(\vec{u},\vec{v}')$ of $(M^{(\vec{u},\vec{v})})_{\vec{Y} \gets \vec{y}}$. 

We can also evaluate formulas with respect to a partial causal setting $(M,\vec{u})$, or with respect to a model $M$. For basic causal formulas, $(M,\vec{u}) \sat  [\vec{Y} \gets \vec{y}]\phi$ iff $(M,\vec{u},\vec{v}) \sat  [\vec{Y} \gets \vec{y}]\phi$ holds for all states $\vec{v}$ such that $(\vec{u},\vec{v})$ is a solution of $M$. In a similar fasion, we define that $M \sat  [\vec{Y} \gets \vec{y}]\phi$ iff $(M,\vec{u},\vec{v}) \sat  [\vec{Y} \gets \vec{y}]\phi$ holds for all solutions $(\vec{u},\vec{v})$ of $M$. %, and similarly for $(M,\vec{v}) \sat \psi$ with the roles of $\vec{u}$ and $\vec{v}$ reversed. %Note: think about this more: (The generalization to subtuples $\vec{X} \subseteq \U \cup \V$ and settings $(M,\vec{x}) \sat [\vec{Y} \gets \vec{y}]\phi$  is straightforward.)
We again inductively extend to causal formulas in the standard way: $(M,\vec{u}) \sat [\vec{Y} \gets \vec{y}]\phi_1 \land [\vec{Z} \gets \vec{z}]\phi_2$ iff $(M,\vec{u}) \sat [\vec{Y} \gets \vec{y}]\phi_1$ and $(M,\vec{u}) \sat [\vec{Z} \gets \vec{z}]\phi_2$, and similarly for $\lnot $ and $\lor$. Likewise for $M \sat \psi$. 

We have now defined three different semantics for $\cal{L}(\cal{S})$: the first with respect to full causal settings $(M,\vec{u},\vec{v})$, the second with respect to partial causal settings $(M,\vec{u})$, and the third with respect to $M$. I refer the reader to \citep{beckers25} for a discussion and comparison of these semantics.

%%%%%%
\section{Informal Summary and Motivation}\label{sec:inf}

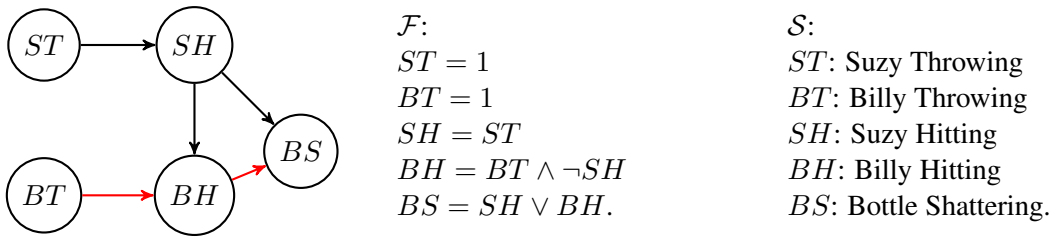
\begin{figure}
\begin{multicols}{3}
\begin{tikzpicture}[node distance={20mm}, thick, main/.style = {draw, circle}] 
\node[main] (1) {$ST$}; 
\node[main] (2) [below of=1] {$BT$};
\node[main] (3) [right of=1] {$SH$};
\node[main] (4) [right of=2] {$BH$};
\node[main] (5) [below right of=3] {$BS$};

\draw[->,>=stealth',auto] (1) -- (3);
\draw[->,>=stealth',auto,red] (2) -- (4);
\draw[->,>=stealth',auto] (3) -- (4);
\draw[->,>=stealth',auto] (3) -- (5);
\draw[->,>=stealth',auto,red] (4) -- (5);
\end{tikzpicture}
\columnbreak

$\F$:\\
$ST=1$\\
$BT=1$\\
 $SH = ST$\\
 $BH=BT \land \lnot SH$\\
 $BS = SH \lor BH$.
 \columnbreak
 
${\cal S}$:\\
$ST$: Suzy Throwing\\
$BT$: Billy Throwing\\
 $SH$: Suzy Hitting\\
 $BH$: Billy Hitting\\
 $BS$: Bottle Shattering.
\end{multicols}
\caption{$M$ for Example \ref{ex:lp} (Removing the red edges results in structural simplifications of $M^{\vec{v}}$.)}\label{fig:lp}
\end{figure}

In order to get a sense of where we are heading, I here present the informal idea behind the new definition of actual causation that will be developed. Say we observe two events $X=x$ and $Y=y$. The most popular approach to defining what it means for $X=x$ to be an actual cause of $Y=y$ relative to a causal setting $(M,\vec{u},\vec{v})$ is to construct it as a more nuanced version of {\em counterfactual dependence}: if $X$ had been $x'$ instead of $x$, then $Y$ might have been $y'$ instead of $y$, for some $x' \neq x$ and $y' \neq y$. The reason that we need to add nuance is that there exist compelling examples illustrating that counterfactual dependence is not necessary for actual causation (despite it arguably being sufficient). Preemption cases are by far the most famous such examples, and therefore I use a preemption classic that is discussed by \cite{halpernbook} at length to illustrate the problem.
\begin{example}[{\bf Late Preemption}]\label{ex:lp}
Suzy and Billy both throw a rock at a bottle. Suzy's rock gets there first, shattering the bottle. However Billy's throw was also accurate, and would have shattered the bottle had it not been preempted by Suzy's throw. (See Figure \ref{fig:lp}.)
\end{example}

Intuitively, Suzy's throw caused the bottle to shatter and Billy's throw did not. Yet if Suzy had not thrown, the bottle would have shattered nonetheless, and thus there is no counterfactual dependence. Note that Billy's throw serves as a backup process here, and if it weren't for that backup process, then there would be counterfactual dependence. So a natural idea to pursue for defining actual causation is to consider whether there would be counterfactual dependence if the backup process were somehow ignored. To implement this idea requires answering two questions: what exactly does it mean to ``ignore'' the backup process, and why should we be at all interested in considering counterfactual dependence in such a hypothetical situation of ignorance, given that the causal model is as it is? So far the debate on defining actual causation has centered exclusively around the various technical answers to the first question, and almost no attention has been paid to the second question. (Two notable exceptions are \citep{hitchcock17,beckers22a}.) Instead, I propose that answering the second question is what should guide us towards answering the first, for it is only by understanding what {\em function} the notion of actual causation fulfils that we can get a handle on formally defining it. In fact, as we will see, addressing the second question offers insights into the logic of causal discovery that are of interest quite aside from the issue of actual causation.

Coming back to our example, what if we are not fully confident about all details of the causal model, but rather, the model as we outlined it is merely the simplest model that is consistent with all the available observational and experimental data? Erring on the side of caution, in  that case we should consider the possibility that the real model allows for a richer set of data. In particular, as it turns out, we can systematically increase the set of data consistent with the model by simplifying the structure of the model, and thus any such simplification represents a model that we still consider plausible. It's just that the more we simplify the structure the less plausible the model becomes.

More specifically, given a causal setting $(M,\vec{u},\vec{v})$, we should extend our interest from relations of counterfactual dependence relative to $(M,\vec{u},\vec{v})$ to such relations relative to any $(M^{simp},\vec{u},\vec{v})$, where $M^{simp}$ is an appropriate simplification of $M$. (As we will see in Section \ref{sec:cd}, that relations of counterfactual dependence are themselves of interest is not hard to show.) I claim that actual causation consists precisely of all relations of counterfactual dependence in such simplifications. 

\iffalse
 the models that are constructed out of the original model by making them more complex. 

merely the default model that we have hypothesized as the most complex model we consider to be plausible? In that case, our interest should extend not just to the model as is, but also to any simplification of that model, for any such simplification represents a model that we consider plausible, albeit less plausible than the default model. 

More specifically, given a causal setting $(M,\vec{u},\vec{v})$, we should extend our interest from relations of counterfactual dependence relative to $(M,\vec{u},\vec{v})$ to such relations relative to any $(M^{simp},\vec{u},\vec{v})$, where $M^{simp}$ is a simplification of $M$. (As we will see in Section \ref{sec:cd}, that relations of counterfactual dependence are an efficient means of offering useful information about a model is not hard to show.) I suggest that actual causation consists precisely of all relations of counterfactual dependence in such simplifications. 

Informally, these simplifications are the result of removing edges from $\G_M$ in such a way that 
\begin{itemize}
\item each edge removal is also a removal of an ancestor-descendant pair;
\item no cases of counterfactual dependence with respect to $(\vec{u},\vec{v})$ are removed;
\item the increase in cases of counterfactual dependence is minimal.
\end{itemize}
\fi

Informally, these simplifications are the result of removing edges from $\G_M$ in such a way that (1) each edge removal is also a removal of an ancestor-descendant pair; (2) no cases of counterfactual dependence w.r.t. $(\vec{u},\vec{v})$ are removed; and (3) the increase in cases of counterfactual dependence w.r.t. $(\vec{u},\vec{v})$ is minimal with respect to the resulting graph.
%it does not remove any case of counterfactual dependence with respect to $(\vec{u},\vec{v})$, and results only in a minimal increase of such cases, thereby guaranteeing that any novel such case is solely the result of simplifying the model. 

As it turns out, in Example \ref{ex:lp} the only edges that can be so removed are those in red. In this case, ignoring the backup process coincides with removing the edge $BH \rightarrow BS$, which results in the equation for $BS$ taking on the nondeterministic form: $BS =1$ if $SH=1$ else $BS \in {\{0,1\}}$. (Roughly: $BH$'s role in the equation is replaced with nondeterminism.) According to that equation, we get that if Suzy had not thrown, then the bottle might have not shattered, and hence Suzy's throw counts as an actual cause of the shattering. We revisit this example in detail in Section \ref{sec:ac}.
%%%%%

\section{Counterfactual Dependence}\label{sec:cd}

At the heart of the interventionist approach to causation that was mathematically developed by \cite{pearl:book2} and philosophically developed by \cite{woodward} lies the general idea that for $X$ to be a cause of $Y$ means for $X$ to be a tool that allows one a certain amount of control over $Y$: in some situation, interventions on $X$ from one value to another possibly result in changes to the value of $Y$. In other words, $X$ is a cause of $Y$ if in some situation $Y$ depends on $X$, where dependence is to be understood in terms of interventions. {\em Counterfactual dependence} amounts to making this idea precise when considering a particular, actual, situation, in which the causal mechanisms have run their course and produced particular values for all the variables. Within the context of causal models, counterfactual dependence is thus a statement about what is the case in an actual world $(\vec{u},\vec{v})$ that is governed by a causal model $M$, i.e., it is relative to a causal setting $(M,\vec{u},\vec{v})$. 

I first present the standard definition of counterfactual dependence as it appears in \cite{halpernbook}, where it was defined for deterministic causal models, meaning that we are in a situation of full control: intervening to change $X$ from its actual value to a specific other value results in {\em certainly} changing the value of $Y$. In the deterministic case a partial causal setting $(M,\vec{u})$ has only one solution $(\vec{u},\vec{v})$, and therefore it can be identified with the full causal setting $(M,\vec{u},\vec{v})$. We label this notion {\em dependence*} to indicate its preliminary nature. (We assume throughout that $X \neq Y$.)

\begin{definition}\label{def:CD1}
$Y = y$ \emph{depends*} on $X=x$ in 
$(M,\vec{u})$ if there exists $x' \in \R(X)$ s.t.:

{\rm [CD1.]} $(M,\vec{u}) \sat \<\>(X = x  \land Y=y)$ and {\rm [CD2.]} $(M,\vec{u}) \sat [X \gets x'] Y \neq y.$\end{definition}
\iffalse
\begin{definition}\label{def:CD1}
$Y = y$ \emph{depends*} on $X=x$ in 
$(M,\vec{u})$ if there exists $x' \in \R(X)$ s.t.:
\begin{description}
\item[{\rm CD1.}]\label{cd11} $(M,\vec{u}) \sat \<\>(X = x  \land Y=y)$. 
\item[{\rm CD2.}]\label{cd21} $(M,\vec{u}) \sat [X \gets x'] Y \neq y.$\end{description}\end{definition}\fi
In the deterministic case, Def. \ref{def:CD1} states informally that: (CD1) $X=x$ and $Y=y$ are the values of $X$ and $Y$ in the actual world, and (CD2) if $X$ had been $x'$, then $Y$ would certainly have taken on a non-actual value. In the nondeterministic case, Def. \ref{def:CD1} states that {\em possibly}, there is  dependence, in the sense just described. Indeed, we have the following trivial proposition. 
\pro If $Y = y$ depends* on $X=x$ in $(M,\vec{u})$ then there exists some $\vec{v} \in \R(\V)$ and $x' \in \R(X)$ s.t. $(M,\vec{u},\vec{v}) \sat (X = x  \land Y=y)$ and $(M,\vec{u},\vec{v}) \sat [X \gets x'] Y \neq y$.\epro
The idea of dependence that we started out with above, however, merely requires the {\em possibility} of changing the value of $Y$. Therefore the generalization of counterfactual dependence to nondeterministic models is as follows.

\begin{definition}\label{def:CD2}
$Y = y$ \emph{counterfactually depends*} on $X=x$ in 
$(M,\vec{u},\vec{v})$ if there exists $x' \in \R(X)$ s.t.:%such that the following two conditions hold: 

{\rm [CD1.]} $(M,\vec{u},\vec{v}) \sat (X = x  \land Y=y)$ and {\rm [CD2.]} $(M,\vec{u},\vec{v}) \sat  \langle X \gets x'\rangle Y \neq y.$
\iffalse
\begin{description}
\item[{\rm CD1.}] $(M,\vec{u},\vec{v}) \sat (X = x  \land Y=y)$. 
\item[{\rm CD2.}]  $(M,\vec{u},\vec{v}) \sat  \langle X \gets x'\rangle Y \neq y.$\end{description}
\fi
\end{definition}
In the acyclic deterministic case, Definitions \ref{def:CD1} and \ref{def:CD2} are easily seen to be equivalent, whereas in the acyclic nondeterministic case, Definition \ref{def:CD1} implies Definition \ref{def:CD2} (for some $\vec{v}$) but not the reverse.

Both of these definitions can be understood as capturing {\em observational dependence}, in the sense that they consist of dependence relative to an {\em observed} causal model $M$. Put differently, the actual world $(\vec{u},\vec{v})$ belongs to the observational rung of Pearl's three-fold causal hierarchy \citep{bareinboim22}. We can also consider dependence in a world that is brought about {\em after} $M$ has been intervened upon, so that the causal model takes on the form $M_{\vec{Z} \gets \vec{z}}$. (For example, we can imagine that the actual world takes place in an experimental set-up in which some variables are being held fixed.) Except for the fact that some of its equations are constant equations, a causal model $M_{\vec{Z} \gets \vec{z}}$ is just that, a causal model. Hence we can generalize our definitions to include them.%this wider class of causal models. 

\subsection{Interventional Dependence}\label{sec:id}

\begin{definition}\label{def:CD3}
$Y = y$ \emph{depends} on $X=x$ in 
$(M_{\vec{Z} \gets \vec{z}},\vec{u})$ if $X \not \in \vec{Z}$ and there exists $x' \in \R(X)$ s.t.:

{\rm [CD1.]} $(M_{\vec{Z} \gets \vec{z}},\vec{u}) \sat \<\>(X =x  \land Y=y)$ and {\rm [CD2.]} $(M_{\vec{Z} \gets \vec{z}},\vec{u}) \sat [X \gets x'] Y \neq y.$
\iffalse
\begin{description}
\item[{\rm CD1.}]\label{cd13} $(M_{\vec{Z} \gets \vec{z}},\vec{u}) \sat \<\>(X =x  \land Y=y)$. 
\item[{\rm CD2.}]\label{cd23} $(M_{\vec{Z} \gets \vec{z}},\vec{u}) \sat [X \gets x'] Y \neq y.$\end{description}
\fi
\end{definition}

\begin{definition}\label{def:CD4}
$Y = y$ \emph{counterfactually depends} on $X=x$ in 
$(M_{\vec{Z} \gets \vec{z}},\vec{u},\vec{v})$ if $X \not \in \vec{Z}$ and there exists  $x' \in \R(X)$ s.t.:

{\rm [CD1.]} $(M_{\vec{Z} \gets \vec{z}},\vec{u},\vec{v}) \sat (X = x  \land Y=y)$ and {\rm [CD2.]} $(M_{\vec{Z} \gets \vec{z}},\vec{u},\vec{v}) \sat  \langle X \gets x'\rangle Y \neq y.$
\iffalse
\begin{description}
\item[{\rm CD1.}] $(M_{\vec{Z} \gets \vec{z}},\vec{u},\vec{v}) \sat (X = x  \land Y=y)$. 
\item[{\rm CD2.}]  $(M_{\vec{Z} \gets \vec{z}},\vec{u},\vec{v}) \sat  \langle X \gets x'\rangle Y \neq y.$
\end{description}
\fi

We say that $Y$ {\em depends on} $X$ if there exists some setting $(M_{\vec{Z} \gets \vec{z}},\vec{u},\vec{v})$ and some values such that $Y=y$ counterfactually depends on $X=x$. 

We say that $Y$ {\em directly depends on} $X$ if  $Y$ depends on $X$ for $\vec{Z} =\V - \{X,Y\}$.
\end{definition}
In Definitions \ref{def:CD3} and \ref{def:CD4} the actual world $(\vec{u},\vec{v})$ belongs to the interventional rung of Pearl's causal hierarchy: it represents a world that is observed after an intervention took place. For all variants of dependence, the formula in CD2 belongs to the counterfactual rung, in the sense that it is a statement about worlds that are counterfactual relative to $(\vec{u},\vec{v})$. As with our original notions of dependence, it easily follows that Definition \ref{def:CD3} implies the existence of some $\vec{v}$ such that Definition \ref{def:CD4} holds, and thus also that if $Y=y$ depends on $X=x$ for some $(M_{\vec{Z} \gets \vec{z}},\vec{u})$ then $Y$ depends on $X$. 

Statements of dependence matter because they are highly informative for learning the graph $\G_M$ of a causal model. 

\thm\label{thm:det} Given a causal model $M$ and $X, Y \in \V$, if $Y$ depends on $X$ then $X$ is an ancestor of $Y$ in $\G_M$.
\ethm

\prf  Proofs of all Theorems are to be found in the Appendix. \eprf

The reverse direction does not hold because the ancestor relation is transitive (per definition), whereas dependence is not. (Here is a simple counterexample. $Y= abs(X)$, $X=A$ if $Z=1$ and $X=-A$ if $Z=0$. Then $Z$ is an ancestor of $Y$, but $Y$ does not depend on $Z$.) If we restrict ourselves to the direct case, the two notions {\em do} coincide.

\thm\label{thm:dirdet} Given a causal model $M$ and $X, Y \in \V$, $X \rightarrow Y \in \G_M$ iff $Y$ directly depends on $X$.
\ethm

These results help to shed light on the importance of statements of counterfactual dependence in both scientific practice and in everyday life: they offer a concise and efficient means of communicating the causal structure of a model. We explore this insight further in Sections \ref{sec:disc} and \ref{sec:ac}, leading to my proposal for defining actual causation. 

%Added 13/9
\iffalse
[Integrate this, it's a great insight!

How does this definition fare as a definition of AC? It looks even weaker than Original HP, but is it? 

Consider the famous Loader counterexample to the Original HP:
$Y = (X \land D) \lor A$, and a context such that $X=1, A=1, D=0, Y=1$.

Then: $(M,\vec{u} \sat [A \gets 0, D \gets 1, X \gets 0] Y=0$ and $(M,\vec{u}) \sat [A \gets 0, D \gets 1, X \gets 1] Y=1$. 

But under this ``intervention first'' interpretation, we actually have: X=1, A=0, D=1. And obviously X=1 is a cause of Y=1 there!

This opens up a new perspective (justification?) of the Modified HP definition: if $\vec{z}$ occur ``naturally'', i.e., they agree with the equations even without intervening, then we cannot distinguish $M$ and $M_{\vec{Z} \gets \vec{z}}$. ]
\fi
%%%%

Interventional dependence opens up an interesting conceptual distinction that -- as far as I am aware -- has been overlooked so far, namely the distinction between $(M_{\vec{Z} \gets \vec{z}},\vec{u},\vec{v}) \sat \<\vec{W} \gets \vec{w}\>\phi$ and $(M,\vec{u},\vec{v'}) \sat \<\vec{W} \gets \vec{w},\vec{Z} \gets \vec{z}\> \phi$. (Throughout the following we assume that $\vec{Z} \cap \vec{W}=\emptyset$.) The reason that it has been overlooked is simple: it cannot be made visible using deterministic models, for there it holds that \begin{equation}\label{eq:int}(M_{\vec{Z} \gets \vec{z}},\vec{u},\vec{v}) \sat \<\vec{W} \gets \vec{w}\>\phi \text{ iff } (M,\vec{u},\vec{v'}) \sat\<\vec{Z} \gets \vec{z},\vec{W} \gets \vec{w}\> \phi.\end{equation}
And yet conceptually these two types of statements are quite different. For sake of simplicity, for now let us assume that $\vec{W}=\emptyset$. Then the LHS of (\ref{eq:int}) states that if the causal model is such that it contains the equations $\F^{\vec{Z} \gets \vec{z}}$ and if this has produced the actual world $(\vec{u},\vec{v})$, then it might be the case that $\phi$. Simply put: intervention first, observation second. The RHS, on the other hand, states that if the causal model is such that it contains the equations $\F$ and if this has produced the actual world $(\vec{u},\vec{v'})$, then if $\vec{Z}$ were $\vec{z}$ it might have been the case that $\phi$. Simply put: observation first, intervention second. So the former is a statement about what is the case in the {\em actual} world (that just happens to have been brought about after an intervention took place) and the latter is a statement about what is the case in a {\em counterfactual} world. The equivalence (\ref{eq:int}) breaks down in the nondeterministic case, as it should. Here is a very simple example. 

\xam\label{ex:two} Imagine two situations sharing identical background conditions $\vec{u}$ that involve the possible recovery $Y$ of a patient who can undergo one of two possible treatments $X$, each of which is sometimes effective. Assume that one treatment ($X=1$) is the default in such situations. In situation 1, the patient is given the deviant treatment ($X \gets 0$), and we observe that the patient does not recover ($Y=0$). In this situation, it plainly holds that the patient certainly does not recover. In situation 2, we observe that the patient receives the default treatment ($X=1$) and does not recover, and we then consider whether they would have recovered if they had received the deviant treatment instead. In this situation, the patient might have recovered. Formally this translates into a model in which the equation for $Y$ is $Y \in \{0,1\}$ for both of the parent values $X=1$ and $X=0$, and the equation for $X$ is $X=1$. Then -- situation 1 -- $(M_{X \gets 0}, \vec{u}, X=0,Y=0) \sat Y=0$ but also -- situation 2 -- $(M,\vec{u},X=1,Y=0) \sat \<X \gets 0 \> Y=1$.
\exam 

%%%
%Added 13/09
Nonetheless, equivalence (\ref{eq:int}) remains also for the nondeterministic case when considering partial settings (which is the case in Definition \ref{def:CD3}). Intuitively, the explanation for this goes as follows. Recall from Section \ref{sec:lan} that causal formulas (relative to the original model $M$) are evaluated after two operations on $M$ have taken place, first refinement and then intervention, expressed by $(M^{(\vec{u},\vec{v})})_{\vec{Z} \gets \vec{z}}$. The refinement operation updates the equations in line with the observation $\V =\vec{v}$, and thus this ordering aligns with the ``observation first, intervention second'' motto. That the equivalence breaks down for full causal settings is because the refinement and intervention operations do not commute: $(M^{(\vec{u},\vec{v})})_{\vec{Z} \gets \vec{z}} \neq (M_{\vec{Z} \gets \vec{z}})^{(\vec{u},\vec{v})}$. Moreover, as a partial causal setting $(M,\vec{u})$ does not contain an observation, we evaluate formulas by considering all possible observations, which effectively neutralizes the refinement operation and thus the issue does not arise. %(see also the result from [Anonymized] showing that the semantics for partial settings for nondeterministic models generalizes the standard semantics for deterministic models).

\thm\label{thm:part} Given a causal model $M$, for any basic causal formula $\phi$ and for any choice of $\vec{Z}, \vec{W} \subseteq \V$, $\vec{z} \in \R(\vec{Z})$, $\vec{w} \in \R(\vec{W})$, and $\vec{u} \in \R(\U)$, we have that 
$$(M_{\vec{Z} \gets \vec{z}},\vec{u}) \sat \<\vec{W} \gets \vec{w}\>\phi \text{ iff } (M,\vec{u}) \sat \<\vec{Z} \gets \vec{z},\vec{W} \gets \vec{w}\> \phi.$$ 
\ethm

%%%

Although overlooked within the context of DSCMs, the distinction between the two types of statement is in fact a very important one within the context of Pearl's {\em probabilistic} treatment of causation. The probabilistic counterpart of the ``intervention-first'' statement is the subject of Pearl's well-known {\em do}-calculus, which offers us rules that allow us to reduce interventional probabilities to probabilities with fewer interventions. Concretely, it is concerned with statements of the form $P_{\vec{Z} \gets \vec{z}}(Y=y | \vec{X}=\vec{x})$, where the subscript $\vec{Z} \gets \vec{z}$ indicates that $P$ is the probability distribution over $\V$ {\em after} the intervention $\vec{Z} \gets \vec{z}$. (As a consequence, the observed condition $\vec{X}=\vec{x}$ is an observation after the intervention, and thus $P_{\vec{Z} \gets \vec{z}}(\vec{X}=\vec{x} | \vec{X}=\vec{x})=1$.) The probabilistic counterpart of the ``observation-first'' statement is the subject of Pearl's probabilities of counterfactuals, which are statements of the form $P(Y_{\vec{Z} \gets \vec{z}}=y | \vec{X}=\vec{x})$. Here the appearance of the subscript $\vec{Z} \gets \vec{z}$ next to $Y$ (instead of $P$) indicates that $P$ is the distribution after {\em first} having observed $\vec{X}=\vec{x}$, and {\em then} applying the intervention $\vec{Z} \gets \vec{z}$. (As a consequence, in general we do not have that $P(\vec{X}_{\vec{Z} \gets \vec{z}}=\vec{x} | \vec{X}=\vec{x})= 1$.) More generally, 
\begin{equation}\label{eq:prob}P_{\vec{Z} \gets \vec{z}}(Y=y | \vec{X}=\vec{x}) \neq P(Y_{\vec{Z} \gets \vec{z}}=y | \vec{X}=\vec{x}).\end{equation} 
So why does a distinction that is fundamental to Pearl's probabilistic framework for causation not have a counterpart in his (and Halpern's) logical framework for causation? Because the former is concerned with the setting in which there is uncertainty over the context $\vec{u}$, whereas the latter is formulated relative to a known $\vec{u}$, and for DSCMs the distinction is visible only in the first of these. It suffices to condition the above probabilities on a known context $\vec{u}$ and then invoke our earlier equivalence to make this clear (under the assumption that $M$ is deterministic):\footnote{Here $1_{\{condition\}}$ denotes the indicator function that returns $1$ if the $condition$ is true and $0$ otherwise.}
\begin{align*}
P_{\vec{Z} \gets \vec{z}}(Y=y | \U=\vec{u}, \V =\vec{v}) = 1_{\{(M_{\vec{Z} \gets \vec{z}},\vec{u},\vec{v}) \sat Y=y\}}
 = 1_{\{(M,\vec{u},\vec{v}') \sat [\vec{Z} \gets \vec{z}]Y=y\}}=P(Y_{\vec{Z} \gets \vec{z}}=y | \U=\vec{u}, \V=\vec{v}').\end{align*}
Since the equivalence breaks down in the nondeterministic case, so do these equalities, and therefore the probabilistic and logical treatments of causation are more closely aligned in the nondeterministic case than they are in the standard deterministic case. 

Still, even in the deterministic case interventional dependence opens up a novel subject of investigation, namely the probabilistic counterpart to interventional dependence that we arrive at when we drop our earlier assumption that $\vec{W}=\emptyset$. In that case, the LHS of (\ref{eq:int}) represents a situation in which we have ``intervention first, observation second, intervention third'', which is a {\em counterfactual} statement relative to an {\em actual world} that was brought about {\em after an intervention}. Sticking to the notation used above, the probabilistic counterpart of such statements would be $P_{\vec{Z} \gets \vec{z}}(Y_{\vec{W} \gets \vec{w}}=y | \vec{X}=\vec{x})$. So although such statements are perfectly sensible within Pearl's existing framework, Pearl does not discuss them. 

I can only speculate, but I suspect the reason for this oversight lies in Pearl's inconsistent use of notation throughout his book \citep{pearl:book2}. Within the context of the {\em do}-calculus (p. 67), he equates the following notation: $P_{Z \gets z}(Y=y) = P(Y=y | {\hat Z}={\hat z}) = P(Y=y | do(Z=z))$. This seems to imply that $P( . | do( . ))$ is his notation for the ``intervention-first, observation-second'' type of statement. Yet within the context of counterfactuals (p. 221 and p. 232), he equates $P(Y=y | do(Z=z))$ with $P(Y_{Z \gets z} = y)$, implying instead that $P( . | do( . ))$ is his notation for the ``observation-first, intervention-second'' type of statement. This ambiguity is harmless in two extreme cases: the case in which we do not condition on any observation, or the case in which we condition on a complete context $\vec{u}$ (in the deterministic case, that is). Outside of these extremes, however, inequality (\ref{eq:prob}) shows that this ambiguity results in a contradiction.\footnote{Interestingly, in the last chapter Pearl replies to comments by readers. In one comment (p. 354) a reader points out that Pearl lacks the notation to distinguish between an observation that occurs before, and one that occurs after, an intervention. In his reply, Pearl claims that he does make such a distinction in the chapter on counterfactuals. Yet such a distinction is nowhere to be found, further confirming that Pearl was overlooking the nuances here discussed.}

\iffalse 
Here is a very simple example. Imagine two situations that share identical background conditions $\vec{u}$ and in which we observe a coin-flip ($Y$) and some entirely unrelated event ($X$). In situation 1, we intervene to set $X$ to $x$, and then observe that the coin lands heads. In this situation, it plainly holds that the coin lands heads. In situation 2, we observe that the coin lands tails and $X=x'$, and then we consider the outcome of the coin-flip if $X$ were $x$ instead. In this situation, the coin would still have landed tails, for it is not in any way determined by $X$. Formally: consider a model in which the equation for $Y$ is $Y \in \{0,1\}$ and there is some other variable $X$. Then $(M_{X \gets x}, \vec{u}, X=x,Y=1) \sat Y=1$ but $(M,\vec{u},X=x',Y=0) \sat [X \gets x] Y=0$.
\fi

\section{The Logic of Causal Discovery}\label{sec:disc}

We now take advantage of Theorems \ref{thm:det}, \ref{thm:dirdet}, and \ref{thm:part}, to move on to considering the process of causal discovery of an NSCM $M$ from an idealized logical perspective. Although obviously extremely important, we here set aside entirely all statistical issues justifying the inductive leap from a finite sample to a general theory, as well as all issues regarding the impracticality of performing perfect interventions, and criteria for inferring that two observations occurred under identical conditions. Instead, we start from the idealized assumption that through rigorous experimentation and well-justified assumptions about the background conditions that make up a context $\vec{u}$, we have learned a set of ``interventionist statements of possibility'' that hold in the ground truth model $M$, i.e., we have learned a set of statements $S$ of the form $(M_{\vec{X} \gets \vec{x}},\vec{u}) \sat \<\> \V=\vec{v}$, or equivalently (Thm. \ref{thm:part}), $(M,\vec{u}) \sat \langle \vec{X} \gets \vec{x}\rangle \V=\vec{v}$, as well as the maximal set of consequences that can be derived from such statements (through the axiomatization given in \citep{beckers25}). Furthermore, assume that $S$ contains at least one element for each combination of a context $\vec{u}$ and an intervention $\vec{X} \gets \vec{x}$, so that we can imagine having performed each possible experiment in each possible context at least once. 
Per assumption, if $(M,\vec{u}) \sat \langle \vec{X} \gets \vec{x}\rangle \phi \in S$, then $(M,\vec{u}) \sat \langle \vec{X} \gets \vec{x}\rangle \phi$. In order to fully identify $M$, we need at least the further assumption that our sample of experiments was sufficiently large for $S$ to contain {\em all} possibilities, so that if $(M,\vec{u}) \sat \langle \vec{X} \gets \vec{x}\rangle \phi \not \in S$ then  $(M,\vec{u}) \sat [\vec{X} \gets \vec{x}] \lnot \phi$. Let us call this the {\bf exhaustivity assumption}. 

Under the {\bf exhaustivity assumption} such a set $S$ gets us quite a way towards identifying $\G_M$: $S$ contains all statements of dependence relative to partial settings (Def. \ref{def:CD3}), both direct and not direct, so using Theorem \ref{thm:dirdet} this allows us to identify a partial graph $\G_S \subseteq \G_M$. However, completely identifying $\G_M$ requires knowledge of all statements of {\em counterfactual} dependence (Def. \ref{def:CD4}), and those are not exhausted by  dependence relative to partial settings. Example \ref{ex:two} offers an illustration: we cannot identify the edge $X \rightarrow Y \in \G_M$ from statements of dependence relative to partial settings alone, for those are equally compatible with the equation for $Y$ simply being $Y \in \{{0,1}\}$, so that $X \not \rightarrow Y \in \G_M$. Only upon learning a counterfactual statement such as $(M,\vec{u},X=1,Y=0) \sat \<X \gets 0 \> Y=1$ can we rule out the latter. In fact, $\G_S \subseteq \G_M$ is {\em exactly} what we learn about the graph from $S$ and {\bf exhaustivity}, in the sense that we cannot rule out any acyclic supergraph of $\G_S$ as the complete graph $\G_M$. Concretely, as the following result shows, the problem of identifying the model $M$ reduces to the problem of identifying those edges in $\G_M$ that go beyond $\G_S$.

%Note: the only edges that are not identifiable from just interventionist statements, are those such that for each setting $\vec{z}$ of the other parents of $Y$, the possible values for $Y$ are identical across all values of $X$, and for at least one $\vec{z}$ the possible values of $Y$ are more than one for all values of $X$. So exactly the kind of situation in Example 1. 

\thm\label{thm:exh} Given a set of interventionist statements $S$ as described above and an acyclic graph $\G_M \supseteq \G_S$ over the same signature ${\cal S}$, the following construction uniquely defines a causal model $M = ({\cal S},\F,\G_M)$ such that for each $\vec{u} \in \R(\U)$, $\vec{X} \subseteq \V$, $\vec{x} \in \R(\vec{X})$, and basic formula $\phi \in {\cal L}({\cal S})$: $$(M,\vec{u}) \sat \langle \vec{X} \gets \vec{x}\rangle \phi \in S \text{ iff } (M,\vec{u}) \sat \<\vec{X} \gets \vec{x} \> \phi.$$
For all $X \in \V$, $x \in \R(X)$, $\vec{a} \in \R(\vec{Pa_X} \cap \V)$ and $\vec{b} \in \R(\vec{Pa_X} \cap \U)$, we have that $x \in f_X(\vec{a},\vec{b})$ iff $(M,\vec{u}) \sat \langle\vec{A} \gets \vec{a} \rangle X=x \in S$ for some $\vec{u} \in \R(\U)$ such that $\vec{u}_{\vec{B}}=\vec{b}$.

We let $M_S$ denote the model corresponding to $\G_S$, and call it the {\em \bf default model} for $S$.\ethm

We do get unique identifiability of $M$ just from a set $S$ if we assume that it is a {\em deterministic} SCM, since in that case dependence and counterfactual dependence coincide and thus $\G_S=\G_M$. This allows us to reconceptualize the assumption of determinism as simply a more extreme version of the {\bf exhaustivity assumption}: it amounts to the assumption that exhaustivity is reached after doing each experiment in each context only once, because it is assumed that each context and intervention results in a {\em unique} solution. As a consequence, under the assumption of determinism, just a single instance of observing two solutions for the same experiment in the same context forces us to conclude that we {\em must} have somehow made a mistake in our experimental setup or our observations. NSCMs allow us to avoid that conclusion. (Indeed, it was the restrictiveness of the uniqueness property that motivated me in \citep{beckers25} to develop NSCMs in the first place.)

Still, even without assuming determinism, {\bf exhaustivity} is a strong assumption, and learning just a single novel interventionist possibility results in its violation. If we just drop it, we are unable to even rule out the extreme model $M$ that allows for {\em all} possibilities, meaning the completely disconnected model in which each variable $X \in \V$ takes on the equation $X \in \R(X)$. Nor are we able to rule out any edge $X \rightarrow Y \in \G_M$. Therefore it would be good to replace it with an assumption that takes {\bf exhaustivity} as the default, but is compatible with it being violated.  To do so, consider that the underlying motivation for {\bf exhaustivity} is in fact an appeal to a certain kind of simplicity assumption, namely the assumption that the best explanation for the data is the simplest model that is able to explain it. Such appeals to the best explanation are standard within scientific methodology, as is the view that simplicity is one of the most important indicators of good explanations. (Philosophers of science call this ``inference to the best explanation'', but among scientists it is more often referred to as ``abduction'' \citep{sep:abduction}.)

Simplicity in this sense corresponds to the set-inclusion partial order for sets of interventionist possibilities: all else being equal, $A$ is simpler than $B$ iff $A \subseteq B$. Let us call this {\em int-set simplicity}. However, the complexity of causal models is also characterized by their causal structure, which is encoded in the graph $\G_M$. There exist many measures of relative graph complexity, but within the context of causal models we require a measure that focuses on the relations of causal dependence. In light of Theorem \ref{thm:det}, I suggest using the ancestor-descendant relation as a measure of simplicity: all else being equal, $\G_A$ is simpler than $\G_B$ iff $\G_A$ is constructed out of $\G_B$ by removing ancestor-descendant pairs. (I explain such constructions in Definition \ref{def:graph} below.)
%$\G_A = \G_B - Anc(R)$, where $Anc(R)$ is some set of ancestor-descendant pairs and the  operation $-$ of removing such pairs from a graph is explained below. 
Let us call this {\em structural simplicity}. Since both measures of simplicity pull in different directions, the most conservative approach is to only rule out those models that are dominated along the combination of both measures, meaning that there exists an alternative model explaining $S$ which is strictly simpler along one measure and is at least as simple along the other. Furthermore, in line with -- but much weaker than -- {\bf exhaustivity}, we assume a preference for int-set simplicity, so that models which are structurally simpler but less int-set simple become evermore abnormal as edges are removed. Let us call all of this the {\bf simplicity assumption}. 

\iffalse
A first consequence is that under {\bf simplicity}, we can use Theorem \ref{thm:exh} to derive that $\G_M \subseteq \G_S$, 

[TO DO: no longer correct...]

since all of the models there constructed are int-set minimal anyway: they imply all of $S$ -- which is a minimal requirement -- and nothing more. That is why we called $M_S$ the default model: under {\bf simplicity} it is the unique int-set minimal model. Nevertheless, there do exist models consistent with {\bf simplicity} that are {\em structurally} simpler than $M_S$. Concretely, for any graph $\G_{T}$ constructed out of $\G_S$, such a corresponding model $M_T$ has to contain a minimal set of interventionist possibilities extending $S$. We work our way up to identifying these models as the {\em structural simplifications} of the model $M_S$.
\fi

Under {\bf simplicity} the most normal models are the structurally simplest, int-set minimal, models. All the models identified in Theorem \ref{thm:exh} are int-set minimal, but $M_S$ is the only one that does not contain any redundant edges, and that is why I take it to be the {\em default} model. There do, however, also exist less normal models consistent with {\bf simplicity}, and those will be structurally simpler than $M_S$. Concretely, for any graph $\G_{T}$ constructed out of $\G_S$, such a corresponding model $M_T$ has to contain a minimal set of interventionist possibilities extending $S$. We work our way up to identifying these models as the {\em structural simplifications} of the model $M_S$.

\subsection{Structural Simplifications}\label{sec:ss}

Structurally simplifying a graph comes down to removing edges in order to remove some ancestor-descendant pairs. Although very similar to removing edges simpliciter, things are complicated slightly by the fact that the removal of an edge $X \rightarrow Y$ does not necessarily result in the removal of an ancestor-descendant pair, for there might remain an indirect path from $X$ to $Y$. (Consider the edge $SH \rightarrow BS$ in Figure \ref{fig:lp} for an illustration.) As the removal of an edge from $\G_S$ comes at the valuable expense of increasing the int-set complexity, we only consider subgraphs in which the removal of edges was worth the price (i.e., it bought us the removal of ancestor-descendant pairs). 

\begin{definition}\label{def:graph} Given graphs $\G_1$ and $\G_2$, we say that $\G_2$ is a {\em structural simplification} of $\G_1$ if $\G_2 \subseteq \G_1$, $Anc(\G_2) \subseteq Anc(\G_1)$, and for any $X \rightarrow Y \in (\G_1 - \G_2)$: $(X,Y) \not \in Anc(\G_2)$.
%$Anc(\G_2 + \{X \rightarrow Y\}) \neq Anc(\G_2)$.
\end{definition}

Int-set simplicity is assessed by considering whether one model {\em interventionally extends} another.

\begin{definition} Given models $M_1$ and $M_2$ over a signature ${\cal S}=(\U, \V,\R)$, we say that $M_2$ is an {\em interventional extension} of $M_1$ if for all $\vec{u} \in \R(\U)$, $\vec{X} \subseteq \V$, $\vec{x} \in \R(\vec{X})$, and basic formulas $\phi \in {\cal L}({\cal S})$, if $(M_1,\vec{u}) \sat \langle \vec{X} \gets \vec{x} \rangle \phi$ then $(M_2,\vec{u}) \sat \langle \vec{X} \gets \vec{x} \rangle \phi.$\end{definition}

\begin{definition} Given a function $f: \R(\vec{X}) \rightarrow \P(\R(Y))$, we define the {\em generalized function} $f^{\vec{Z}}$ for any $\vec{Z}$ as $f^{\vec{Z}}: \R(\vec{Z}) \rightarrow \P(\R(Y))$ and  $f^{\vec{Z}}(\vec{z}) := \bigcup\limits_{\{\vec{x} | \vec{x}_{\vec{Z}} = \vec{z}_{\vec{X}}\}} f(\vec{x})$. \end{definition}

Since $\vec{Z}$ can be deduced from the argument $\vec{z}$, we overload notation and write $f(\vec{z})$ instead of $f^{\vec{Z}}(\vec{z})$. Thus we are effectively interpreting $f$ as a family of multi-valued functions with co-domain $\P(\R(Y))$ and different domains. This idea comes in handy when comparing functions whose domains consist of different parents, which is what is required to construct structural simplifications.

\begin{definition}\label{def:simp} Given models $M_1$ and $M_2$ over a signature ${\cal S}=(\U, \V,\R)$, we say that $M_2$ is a {\em structural simplification} of $M_1$ if $\G_{M_2}$ is a structural simplification of $\G_{M_1}$ and for each $X \in \V$ we have that: $f_X(\vec{Pa_X^2})=g_X(\vec{Pa_X^2})$, where $f_X$ and $g_X$ respectively denote $X$'s function in $M_2$ and $M_1$, and $\vec{Pa_X^2}$ are $X$'s parents in $M_2$.
\end{definition}
%paper to cite: https://arxiv.org/abs/2311.18639
%possibly also look into: https://proceedings.neurips.cc/paper_files/paper/2019/file/c3d96fbd5b1b45096ff04c04038fff5d-Paper.pdf

Informally, structurally simplifying a model by removing an edge $Z \rightarrow X$ means to re-interpret $f_X$ so that for any given input of the remaining parents of $X$, we return all values of $X$ that the original $f_X$ could have given for each value of $Z$. (Alternatively, one could implement the same idea by replacing $Z \rightarrow X$ with $Z' \rightarrow X$ for some novel exogenous variable $Z'$ that takes over the role of $Z$ but is independent of it. That is in fact the definition of edge removal used by \cite{janzing13} -- Def 4. -- within the context of measuring causal influence.)

Structural simplifications satisfy our requirement of adding a minimal set of interventionist possibilities to the original model in order to account for the removal of ancestor-descendant pairs:

\thm\label{thm:red1} Given models $M_1, M_2, M_3$ over a signature ${\cal S}$ such that $M_2$ is a structural simplification of $M_1$ and $\G_{M_3} = \G_{M_2}$, we have that $M_2$ is an interventional extension of $M_1$, and either $M_3$ is not an interventional extension of $M_1$ or $M_3$ is an interventional extension of $M_2$.\ethm
  
Hence we conclude that, for a given set of interventionist possibilities $S$, under {\bf simplicity} the possible causal models are identified as all structural simplifications of the default model $M_S$. Note that the extreme, completely disconnected, model mentioned earlier is a structural simplification of any model, and thus we cannot rule it out entirely. Still, as mentioned the degree of structural simplification is also a degree of abnormality, making the extreme model maximally abnormal.

%Therefore there is a trade-off between $S$-simplicity and $\G$-simplicity: simplifying the graph $\G_S$ forces one to add interventionist possibilities not in $S$. The non-dominated models are precisely those which optimize this trade-off. 

\iffalse
Example to use:

Can adding an edge ever increase the interv poss? Yes. Consider the model: $X=1$ and $Y=1$. Now we add the edge $X \rightarrow Y$ by making it: $X=1$, $Y=1$ if $X=1$, $Y \in \{0,1\}$ if $X=0$. Before we had $(1,1)$, $(X \gets 1, 1)$, $(X \gets 0, 1)$, $(1, Y\gets 1)$, $(1, Y \gets 0)$. We still have that, and yet now we also $(X \gets 0, 0)$.
\fi

\section{Actual Causation}\label{sec:ac}

This is as far as we can get with just knowledge of interventionist possibilities. But what if we were to learn just a {\em single} counterfactual possibility? Contrary to the interventionist case, just a single instance can offer vital novel information without requiring any further assumption. For example, learning that $Y$ counterfactually depends on $X$ in some setting $(M,\vec{u},\vec{v})$ offers us a way of using Theorem \ref{thm:det} to establish that $X$ is an ancestor of $Y$. Doing so allows us to discard all structural simplifications of $M_S$ in which $X$ is not an ancestor of $Y$, or if we like, the generally larger set of simplifications in which $Y$ does not counterfactually depend on $X$.

%Note: What if $X \rightarrow Y$ was just one of the invisible edges, such as in Example 1? Then we don't need to discard the models at all: as long as no cycles are created, we could still add the edge. So at least point out that such invisible edges are an issue. But note that this cannot occur if $M_S$ is deterministic.

Note that if $X$ is not ancestor of $Y$ in $\G_S$, this means discarding all of our possible models. In that case, the counterfactual statement was incompatible with the weakest hypothesis that the above discovery process resulted in, namely that the ground truth $M$ is a structural simplification of $M_S$, and thus we have to re-evaluate. I propose that statements of {\em actual causation} between $X$ and $Y$ correspond to the weakest statement regarding the counterfactual dependence of $Y$ on $X$ in $(M,\vec{u},\vec{v})$ such that it can falsify our weakest hypothesis, and that is the statement that $Y$ counterfactually depends on $X$ in {\em some} structural simplification of the ground truth $(M,\vec{u},\vec{v})$. 

\iffalse
Structural simplifications of a model $M$ are simplifications relative to the interventionist possibilities present in the entire model $M$, but we can of course restrict attention also to a single partial setting $(M,\vec{u})$. Moreover, we can also focus on the {\em counterfactual possibilities} instead of the interventionist possibilities, and restrict attention to a single complete setting $(M,\vec{u},\vec{v})$. Structural simplifications relative to $(M,\vec{u})$ are not structural simplifications relative to all $(M,\vec{u},\vec{v})$. 

Simply consider the trivial model $M_1$ with equations $Y=X$ and $X=\{0,1\}$, where both variables are binary. Now we remove the edge from $X$ to $Y$, resulting in the structural simplification $M_2$ with equations $Y \in \{0,1\}$ and $X \in \{0,1\}$. For both $i=1$ and $i=2$ we have $M_i \sat \langle X \gets 1\rangle Y=1$, and $M_1 \sat \langle X \gets 0\rangle Y=0$. Yet, if we consider the causal settings $(M_i,X=1,Y=1)$ for $i=1$ and $i=2$, we have that  $(M_1,X=1,Y=1)\sat [X \gets 0] Y=0$ and yet $(M_2,X=1,Y=1)\sat [X \gets 0] Y=1$.  
\fi

To make this precise, we first define structural simplifications relative to a specific full causal setting $(M,\vec{u},\vec{v})$, as opposed to an entire model $M$ or to a specific partial setting $(M,\vec{u})$. 

\begin{definition}\label{def:cons} Given models $M_1$ and $M_2$ over a signature ${\cal S}=(\U, \V,\R)$, we say that $(M_2,\vec{u},\vec{v})$ is  a {\em structural simplification} of $(M_1,\vec{u},\vec{v})$ if $M_2^{(\vec{u},\vec{v})}$ is a structural simplification of $M_1^{(\vec{u},\vec{v})}$.\end{definition}

Obviously we would like that, in addition, $M_2$ is a structural simplification of $M_1$, and this is not guaranteed to be the case. Simply consider the trivial one-variable model $M_1$ with equation $Y \in \{0,1\}$. Then the model $M_2$ with equation $Y=1$ is not a structural simplification of $M_1$, but $(M_2,Y=1)$ is a structural simplification of $(M_1,Y=1)$. Still, Proposition \ref{pro:ac} shows that this technicality can be ignored when defining actual causation.

\begin{definition}\label{def:AC5}
$X = x$ is an \emph{actual cause} of $Y=y$ in 
$(M_1,\vec{u},\vec{v})$ if there exists a structural simplification $(M_2,\vec{u},\vec{v})$ of $(M_1,\vec{u},\vec{v})$ so that $Y=y$ counterfactually depends on $X=x$ in $(M_2,\vec{u},\vec{v})$.\end{definition}

\pro\label{pro:ac} If $X = x$ is an actual cause of $Y=y$ in $(M_1,\vec{u},\vec{v})$ then there exists a structural simplification $(M_2,\vec{u},\vec{v})$ of $(M_1,\vec{u},\vec{v})$ so that $M_2$ is a structural simplification of $M_1$ and $Y=y$ counterfactually depends on $X=x$ in $(M_2,\vec{u},\vec{v})$.
\epro 

Note that (by applying Theorem \ref{thm:red1} twice, once to $M_2$ and $M_1$ and once more to $M_2^{(\vec{u},\vec{v})}$ and $M_1^{(\vec{u},\vec{v})}$) this means such structural simplifications result in minimal increases -- given $\G_{M_2}$ -- both of interventionist possibilities in $M_1$ and  counterfactual dependence w.r.t. $(M_1,\vec{u},\vec{v})$.

I have here offered a novel definition of actual causation based entirely on the unique function that it can fulfil in learning a causal model. Moreover,  it is the first definition of actual causation that applies equally to DSCMs and NSCMs. Still, for it to be a contender as a definition of {\em actual causation}, as opposed to some other interesting concept, it should offer defensible verdicts of causation across the enormous variety of examples that are discussed in the literature. That this is indeed the case follows from the fact that, when restricted to DSCMs, my definition behaves almost identically to the one I first developed in \citep{beckers21c} and later reformulated in \citep{beckers22a}. Concretely, the new definition agrees with my previous one on {\em all} of the examples I discuss in the former, and therefore I refer the reader to that work for a comprehensive defense of the verdicts reached by my definition. Here I simply illustrate the definition using Example \ref{ex:lp} and a variant of it that is a rare case where my new definition differs from -- and improves upon -- my old definition.

Applying Definition \ref{def:AC5} to Example \ref{ex:lp} (where $\U=\emptyset$) we get that $ST=1$ is an actual cause of $BS=1$ by considering the simplification $M'$ that is the result of removing the edge $BH \rightarrow BS$. As mentioned, this results in the equation for $BS$ becoming $BS =1$ if $SH=1$ else $BS \in {\{0,1\}}$ (because in $M$ we had $\{1\}_{BS} = 1_{SH} \lor \{1,0\}_{BH}$ and $\{0,1\}_{BS} = 0_{SH} \lor \{1,0\}_{BH}$).
Observe that there is indeed counterfactual dependence in $M'$, since $(M',\vec{v}) \sat \<ST \gets 0\>BS=0$. Lastly, we need to verify that $M'^{\vec{v}}$ is a structural simplification of $M^{\vec{v}}$, which reduces to verifying that $f_{BS}^{\vec{v}}(SH)=g_{BS}^{\vec{v}}(SH)$, and that is a direct consequence of the fact that $f_{BS}^{\vec{v}}(SH)=f_{BS}(SH)$, $g_{BS}(SH)=g_{BS}^{\vec{v}}(SH)$, and the first observation (where, as in Definition \ref{def:simp}, $f_{BS}$ and $g_{BS}$ are the respective functions in $M$ and $M'$).
Now I show that $BT=1$ is not an actual cause of $BS=1$. Clearly $BS=1$ does not counterfactually depend on $BT=1$ in the original model. I leave it to the reader to verify that the only other structural simplifications of $M^{\vec{v}}$ are  those in which either $BT \rightarrow BH$ or $BH \rightarrow BS$  (or both) is removed. Since $BT$ is not an ancestor of $BS$ in any of those models, there is no counterfactual dependence either.

\iffalse
This requires considering all possible structural simplifications $M'$ of $M$, but we can obviously restrict ourselves to models in which $BT$ is an ancestor of $BS$. This leaves considering what happens if we remove any combination of the edges in $\{ST \rightarrow SH, SH \rightarrow BH, SH \rightarrow BS\}$. Removing the first edge or not is without consequence, since we are only considering the counterfactual $BT \gets 0$, and thus the fact that $SH=1$ in the actual world guarantees this will remain the case also in any counterfactual world. (In addition, the edge cannot be removed since that would not result in a structural simplification of $(M,\vec{v})$.) Furthermore, as $SH=1$ guarantees $BS=1$ for $f_{BS}$, we have to  remove the third edge to allow for $BS=0$. But this results in $g_{BS}^{\vec{v}}(BH=1)=1$ {\em and}  $g_{BS}^{\vec{v}}(BH=0)=1$, meaning that $BS=0$ is now impossible to obtain.
\fi

Lastly, I discuss a problem for my previous definition that the new definition does not suffer from. Imagine the same setup from Example \ref{ex:lp}, except that this time Suzy does not throw. Intuitively, we would not want to consider her failure to throw ($ST=0$) to be considered a cause of the bottle shattering ($BS=1$), and neither of the above definitions do so. Now imagine that Suzy is not always accurate ($SA$) when throwing. All it takes for my old definition to change its verdict is to make explicit the obvious fact that Suzy would only hit the bottle if she were accurate, so that the equation for $SH$ becomes $SH=ST \land SA$. That seems an undesirable result, and the current definition does not suffer from it. (For sake of completeness I mention that the undesirable result only shows up in case $SA$ is modelled as an endogenous variable, which is arguably not the best choice. %Still, more complicated examples along similar lines can be constructed to show that the definitions can come apart. 
Instead, I would of course recommend modelling the above using the {\em nondeterministic} equation $SH \in \{0,1\}$ if $ST=1$ and $SH=0$ if $ST=0$.)

It is worth pointing out that my definition can be interpreted as vindicating the intuitive but relentlessly discredited idea that actual causation {\em is} counterfactual dependence after all, although this requires a very broad interpretation of possibility. Given that we consider all structural simplifications possible, actual causation amounts to stating that changing $X=x$ possibly might have resulted in changing $Y=y$. The latter statement is precisely how counterfactual dependence is understood in the counterfactual tradition on actual causation, for it builds on the possible world semantics to counterfactuals, both of which can be traced back to the seminal work of \cite{lewis73,lewis73a}.

\section{Conclusion}

Taking advantage of the increased generality offered by nondeterministic causal models, I have developed a {\em functional} account of actual causation: statements of actual causation test the hypothesis that the set of causal models we consider plausible contains the ground truth. This functional approach is a departure from the existing literature that instead focuses on ambiguous examples and inconsistent intuitions, and owes its inspiration to \cite{woodward21}'s longstanding project of offering a functional account of causal reasoning more generally. In future work I aim to extend this account to a probabilistic one by extending the idea of structural simplifications to the {\em probabilistic} causal models I defined in \citep{beckers25}, including Causal Bayesian Networks. 

\iffalse
The development of structural causal models has resulted in immense progress in our ability to formally represent causal reasoning, inference, and discovery. This work continues this development by combining the recent generalization from deterministic to nondeterministic causal models with a {\em functional} account of actual causation, meaning that the focus on ambiguous examples and  inconsistent intuitions is replaced with a focus on the function that statements of actual causation fulfil. This project is part and parcel of \cite{woodward}'s functional account of causality in general, 
\fi
% Acknowledgments---Will not appear in anonymized version
\acks{I thank the reviewers for their insightful reviews. This work was funded by ARO grant: W911NF-22-1-0061.}

\bibliographystyle{spbasic} 
\bibliography{allpapers}

\appendix

\section*{Proofs of Theorems}

\begin{theorem*}{\bf \ref{thm:det}}
Given a causal model $M$ and $X, Y \in \V$, if $Y$ (deterministically) depends on $X$ then $X$ is an ancestor of $Y$ in $\G_M$.
\end{theorem*}

\prf  Assume $Y$ depends on $X$ in $M$. We show that either $X \in \vec{Pa_Y}$, or there exists some $P \in \vec{Pa_Y}$ such that $P$ depends on $X$. The result then follows by straightforward induction. 

If $X \in \vec{Pa_Y}$ then per definition $X \rightarrow Y \in \G_M$. So let us assume $X \not \in \vec{Pa_Y}$. We know that for some $(M_{\vec{Z} \gets \vec{z}},\vec{u},\vec{v})$, $X \not \in \vec{Z}$ and values $x,x',y$ we have:
\begin{itemize}
\item $(M_{\vec{Z} \gets \vec{z}},\vec{u},\vec{v}) \sat (X = x  \land Y=y)$. 
\item $(M_{\vec{Z} \gets \vec{z}},\vec{u},\vec{v}) \sat  \langle X \gets x'\rangle Y \neq y$.
\end{itemize}

$\vec{Z}=\vec{z}$ for all solutions to $M_{\vec{Z} \gets \vec{z}}$ (Axiom of effectiveness), and thus $Y \not \in \vec{Z}$. Therefore by the first condition, $y \in f(\vec{pa_Y})$, where $\vec{pa_Y}=(\vec{u},\vec{v})_{\vec{Pa_Y}}$. Also, by the second condition, there exists some $\vec{v'}$ such that $\vec{v'}_X=x'$ and $y \not \in f^{(\vec{pa_Y},y)}(\vec{pa_Y'})$, where $\vec{pa_Y'}=(\vec{u},\vec{v}')_{\vec{Pa_Y}}$. Combining both claims, we get that $\vec{pa_Y} \neq \vec{pa_Y'}$, and thus there exists some $P \in \vec{Pa_Y}$ such that  $p=\vec{pa_Y}_P \neq \vec{pa_Y'}_P=p'$. Furthermore, $P \in \V$, since $\U=\vec{u}$ across both solutions. So $p=\vec{v}_P$ and $p'=\vec{v'}_P$.

Therefore we have the following two conditions, which imply that $P$ depends on $X$, as had to be shown:
\begin{itemize}
\item $(M_{\vec{Z} \gets \vec{z}},\vec{u},\vec{v}) \sat (X = x  \land P=p)$. 
\item $(M_{\vec{Z} \gets \vec{z}},\vec{u},\vec{v}) \sat  \langle X \gets x'\rangle P \neq p$.
\end{itemize}
\eprf

\begin{theorem*}{\bf \ref{thm:dirdet}} Given a causal model $M$ and $X, Y \in \V$, $X \rightarrow Y \in \G_M$ iff $Y$ directly depends on $X$.
\end{theorem*}

\prf We start with the implication from right to left. Assume $Y$ directly depends on $X$. Then just as in the proof of Theorem \ref{thm:det} we can conclude that either $X \in  \vec{Pa_Y}$ or there exists some $P \in \vec{Pa_Y}$ which changes value in the solutions used in the two conditions, and $P \in \V$, $P \not \in \vec{Z}$. Given that we have direct dependence, also $\vec{Z}=\V- \{X,Y\}$, and thus $P=X$. 

Now the implication from left to right. First let us assume that $f_Y$ is a deterministic function, recalling our convention that in that case, if $X \in \vec{Pa_Y}$ then there exists some setting $\vec{w}$ of all the other parents and values $x,x',y,y'$ such that $y=f_Y(\vec{w},x) \neq f_Y(\vec{w},x')=y'$. Let $\vec{A}=\vec{Pa_Y}_{\V}$ and $\vec{B}=\vec{Pa_Y}_{\U}$. Let $\vec{u} \in \R(\U)$ denote some setting such that $\vec{u}_{\vec{B}}=\vec{w}_{\vec{B}}$, and let $\vec{z} \in \R(\V - \{X,Y\})$ denote some setting such that $\vec{z}_{\vec{A}}=\vec{w}_{\vec{A}}$.

Say $(\vec{u},\vec{v})$ is a solution to the equations in $M_{\vec{Z} \gets \vec{z}}$, and $\vec{v}_X=x''$. If $f_Y(\vec{w},x'') =y=f_Y(\vec{w},x)$, then we have the following two conditions, which is precisely what we had to show:
\begin{itemize}
\item $(M_{\vec{Z} \gets \vec{z}},\vec{u},\vec{v}) \sat (X = x''  \land Y=y)$. 
\item $(M_{\vec{Z} \gets \vec{z}},\vec{u},\vec{v}) \sat  [X \gets x'] Y \neq y$.
\end{itemize}

If $y''=f_Y(\vec{w},x'') \neq y=f_Y(\vec{w},x)$, then the result follows from the following two conditions:
\begin{itemize}
\item $(M_{\vec{Z} \gets \vec{z}},\vec{u},\vec{v}) \sat (X = x''  \land Y=y'')$. 
\item $(M_{\vec{Z} \gets \vec{z}},\vec{u},\vec{v}) \sat  [X \gets x] Y \neq y''$.
\end{itemize}

Second we assume that $f_Y$ is a multi-valued function, meaning that there exists at least one setting $(\vec{w},x)$ such that there exist $y \neq y'$ for which $\{y,y'\} \subseteq f_Y(x,\vec{w})$. Let $\vec{A}$, $\vec{B}$, $\vec{u}$, and $\vec{z}$ be as before. 

Say $(\vec{u},\vec{v})$ is a solution to the equations in $M_{\vec{Z} \gets \vec{z}}$, and $\vec{v}_X=x''$, $\vec{v}_Y=y''$, so: $(M_{\vec{Z} \gets \vec{z}},\vec{u},\vec{v}) \sat (X = x''  \land Y=y'')$. 

If $x'' \neq x$, we get that $(M_{\vec{Z} \gets \vec{z}},\vec{u},\vec{v}) \sat  \<X \gets x\> Y = y \land \<X \gets x\> Y = y'$. Since we cannot have both $y''=y$ and $y''=y'$, the result follows. 

If $x''=x$, we consider two cases. 

First, consider the case such that for all $x' \neq x$, $f_Y(x',\vec{w}) = f_Y(x,\vec{w})$. This means that for all $x'\neq x$, $\{y,y'\} \subseteq f_Y(x',\vec{w})$. As before, since we cannot have both $y''=y$ and $y''=y'$, the result follows from $(M_{\vec{Z} \gets \vec{z}},\vec{u},\vec{v}) \sat  \<X \gets x'\> Y = y \land \<X \gets x'\> Y = y'$.

Second, consider the case where there exists some $x' \neq x$ such that $f_Y(x',\vec{w})\neq f_Y(x,\vec{w})$. If there exists some $y^* \in f_Y(x',\vec{w})$ so that $y^* \not \in f_Y(x,\vec{w})$, then $(M_{\vec{Z} \gets \vec{z}},\vec{u},\vec{v}) \sat  \<X \gets x'\> Y = y^*$ and the result follows. If there exists some $y^* \in f_Y(x,\vec{w})$ so that $y^* \not \in f_Y(x',\vec{w})$, then there must also exist some solution $(\vec{u},\vec{v'})$ to the equations in $M_{\vec{Z} \gets \vec{z}}$ so that $\vec{v'}_X=x$ and $\vec{v}_Y=y^*$.  The result then follows from observing that $(M_{\vec{Z} \gets \vec{z}},\vec{u},\vec{v'}) \sat  \<X \gets x'\> Y \neq y^*$.
\eprf

\begin{theorem*}{\bf \ref{thm:part}} Given a causal model $M$, for any basic causal formula $\phi$ and for any choice of $\vec{Z}, \vec{W} \subseteq \V$, $\vec{z} \in \R(\vec{Z})$, $\vec{w} \in \R(\vec{W})$, and $\vec{u} \in \R(\U)$, we have that 
$$(M_{\vec{Z} \gets \vec{z}},\vec{u}) \sat \<\vec{W} \gets \vec{w}\>\phi \text{ iff } (M,\vec{u}) \sat \<\vec{Z} \gets \vec{z},\vec{W} \gets \vec{w}\> \phi.$$ 
\end{theorem*}

\prf This is a direct consequence of the following result from \citep{beckers25}.

\begin{theorem*} Given a nondeterministic causal model $M$, we have that for all $\vec{Y} \subseteq \V$, for all $\vec{y} \in \R(\vec{Y})$, for all contexts $\vec{u}$, and for all basic formulas $\phi$: 
$$(M,\vec{u}) \sat [\vec{Y} \gets \vec{y}]\phi\text{ iff }\vec{v} \sat \phi\text{ for all states }\vec{v}\text{ such that }(\vec{u},\vec{v})\text{ is a solution of }M_{\vec{Y} \gets \vec{y}}.$$\end{theorem*}

First note that under our assumption $\vec{W} \cap \vec{Z} =\emptyset$, it holds that $(M_{\vec{Z}\gets\vec{z}})_{\vec{W}\gets\vec{w}}=M_{\vec{Z} \gets \vec{z},\vec{W} \gets \vec{w}}$. 

The result then follows from applying this theorem to both sides of the equivalence in Theorem \ref{thm:partA}, giving:

For the left part, we get that $(M_{\vec{Z} \gets \vec{z}},\vec{u}) \sat \<\vec{W} \gets \vec{w}\>\phi$ iff there exists a state $\vec{v}$ such that $\vec{v} \sat \phi$ and $(\vec{u},\vec{v})$ is a solution to $M_{\vec{Z} \gets \vec{z},\vec{W} \gets \vec{w}}$. 

For the right part, we get that $(M,\vec{u}) \sat \<\vec{W} \gets \vec{w},\vec{Z} \gets \vec{z}\> \phi$ iff there exists a state $\vec{v}$ such that $\vec{v} \sat \phi$ and $(\vec{u},\vec{v})$ is a solution to $M_{\vec{Z} \gets \vec{z},\vec{W} \gets \vec{w}}$. 

\eprf

\begin{theorem*}{\bf \ref{thm:exh}} Given a set of interventionist statements $S$ as described above and an acyclic graph $\G_M \supseteq \G_S$ over the same signature ${\cal S}$, the following construction uniquely defines a causal model $M = ({\cal S},\F,\G_M)$ such that for each $\vec{u} \in \R(\U)$, $\vec{X} \subseteq \V$, $\vec{x} \in \R(\vec{X})$, and basic formula $\phi \in {\cal L}({\cal S})$: $$(M,\vec{u}) \sat \langle \vec{X} \gets \vec{x}\rangle \phi \in S \text{ iff } (M,\vec{u}) \sat \<\vec{X} \gets \vec{x} \> \phi.$$
For all $X \in \V$, $x \in \R(X)$, $\vec{a} \in \R(\vec{Pa_X} \cap \V)$ and $\vec{b} \in \R(\vec{Pa_X} \cap \U)$, we have that $x \in f_X(\vec{a},\vec{b})$ iff $(M,\vec{u}) \sat \langle\vec{A} \gets \vec{a} \rangle X=x \in S$ for some $\vec{u} \in \R(\U)$ such that $\vec{u}_{\vec{B}}=\vec{b}$.

We let $M_S$ denote the model corresponding to $\G_S$, and call it the {\em \bf default model} for $S$.\end{theorem*}

\prf Assume we have some $M = ({\cal S},\F,\G_M)$ and a set $S$ as described. Concretely, for each $\vec{u} \in \R(\U)$, $\vec{X} \subseteq \V$, $\vec{x} \in \R(\vec{X})$, there exists at least one $\V=\vec{v}$ such that $(M,\vec{u}) \sat \langle \vec{X} \gets \vec{x}\rangle \V=\vec{v} \in S$. Furthermore, $S$ contains all consequences derived from such statements using the axiomatization $AX^{\bf scc}_{non}$ of the {\bf single context counterfactual} logic from \citep{beckers25}, which is the logic of NSCMs relative to partial settings $(M,\vec{u})$. %Notably, $S$ does not contain any statement of the form $(M,\vec{u}) \sat [\vec{X} \gets \vec{x}] \V=\vec{v}$ for which $\vec{X} \neq \V$. Let $C$ denote the extension of $S$ such that $(M,\vec{u}) \sat [\vec{X} \gets \vec{x}] \phi \in C$ iff $(M,\vec{u}) \sat \<\vec{X} \gets \vec{x} \> \lnot \phi \not \in S$. Then $C$ contains a maximal consistent set of formulas 

%We can uniquely extend $S$ into a maximal consistent set $C$ of statements 

Since, per assumption about $S$, for all $X \in \V$, $\vec{a} \in \R(\vec{Pa_X} \cap \V)$ and $\vec{b} \in \R(\vec{Pa_X} \cap \U)$, there exists at least one statement of the form $(M,\vec{u}) \sat \langle\vec{A} \gets \vec{a} \rangle \V=\vec{v} \in S$ with $\vec{b}=\vec{u}_{\vec{B}}$, we have by axiom D7 that $(M,\vec{u}) \sat \langle\vec{A} \gets \vec{a} \rangle X=\vec{v}_X \in S$ and thus there exists at least one value $x$ so that $x \in f_X(\vec{a},\vec{b})$, and thus the above construction results in a total NSCM. The rest of the proof proceeds very similarly to the completeness proof for $AX^{\bf scc}_{non}$ as it appears in \citep{beckers25}.

$\Rightarrow$: 

We start with the implication from left to right. Consider some $(M,\vec{u}) \sat \langle \vec{X} \gets \vec{x}\rangle \phi \in S$. Given the construction of $S$, this means that there exists some set of statements of the form $(M,\vec{u}) \sat \langle \vec{X} \gets \vec{x}\rangle \V=\vec{v} \in S$ from which the former statement can be derived using $AX^{\bf scc}_{non}$. Therefore the same holds for any NSCM for which that set of statements is true, and thus we can assume without loss of generality that $\phi$ is of the form $\V=\vec{v}$.

We need to show that $(M,\vec{u}) \sat \langle \vec{X} \gets \vec{x}\rangle \V=\vec{v}$, which by the theorem from \citep{beckers25} that we stated in the proof of Theorem \ref{thm:part} reduces to showing that $(\vec{u},\vec{v})$ is a solution of $M_{\vec{X} \gets \vec{x}}$. By axiom D4, $\vec{v}_{\vec{X}}=\vec{x}$. This means we need to show that for each $Y \in \V - \vec{X}$, $y \in f_Y(\vec{pa_Y})$, where $y=\vec{v}_Y$ and $\vec{pa_Y}=(\vec{u},\vec{v})_{\vec{pa_Y}}$. Using the notation from the theorem, let $\vec{pa_Y}=(\vec{a},\vec{b})$. 

Per construction of $M$, we need to show that $(M,\vec{u}) \sat \langle\vec{A} \gets \vec{a} \rangle Y=y \in S$.

By D3(a), we know that $(M,\vec{u}) \sat \langle\vec{C} \gets \vec{c}, \vec{A} \gets \vec{a} \rangle \V=\vec{v} \in S$, where $\vec{C}=\V - (\{Y\} \cup \vec{A})$ and $\vec{c}=\vec{v}_{\vec{C}}$. By D7, we get that $(M,\vec{u}) \sat \langle\vec{C} \gets \vec{c}, \vec{A} \gets \vec{a} \rangle Y=y\in S$.

Let us consider some $\vec{c'} \in \R(\vec{C})$ such that $(M,\vec{u}) \sat \langle\vec{A} \gets \vec{a}, Y \gets y \rangle \vec{C} =\vec{c'} \in S$. (Per assumption about $S$, such $\vec{c'}$ must exist.) If also $(M,\vec{u}) \sat \langle\vec{A} \gets \vec{a}, \vec{C} \gets \vec{c'}  \rangle Y=y \in S$, then by D5 and D7 we get that $(M,\vec{u}) \sat \langle\vec{A} \gets \vec{a} \rangle Y=y \in S$, as required. 

Remains to consider the case where $(M,\vec{u}) \sat \langle\vec{A} \gets \vec{a}, \vec{C} \gets \vec{c'}  \rangle Y=y \not \in S$. We know that for each $C \in \vec{C}$, $C \rightarrow Y \not \in \G_S$. We show by induction that this results in a contradiction. 

Consider some $C \in \vec{C}$. $C \rightarrow Y \not \in \G_S$ implies that for all $c,c' \in \R(C)$, all $y \in \R(Y)$, $\vec{z} \in \R(\vec{Z})$ with $\vec{Z}=\V - \{C,Y\}$, if $(M,\vec{u}) \sat \langle\vec{Z} \gets \vec{z} \rangle(C=c \land Y=y) \in S$ then $(M,\vec{u}) \sat \langle\vec{Z} \gets \vec{z}, C \gets c'  \rangle Y=y \in S$. (To see why, note that due to D3(a) otherwise it must be that $c \neq c'$ and thus $Y$ directly deterministically depends on $C$, and so by Theorem \ref{thm:dirdet} and per construction of $\G_S$, $C \rightarrow Y \in \G_S$.) As a consequence, if $(M,\vec{u}) \sat \langle\vec{Z} \gets \vec{z} \rangle Y=y \in S$ then for all $c' \in \R(C)$, $(M,\vec{u}) \sat \langle\vec{Z} \gets \vec{z}, C \gets c'  \rangle Y=y \in S$. Taking $\vec{D}=\vec{C} -\{C\}$, $\vec{d}=\vec{c}_{\vec{D}}$, and $c=\vec{c}_C$, from the above we know that $(M,\vec{u}) \sat \langle\vec{A} \gets \vec{a},\vec{D} \gets \vec{d}, C \gets c \rangle Y=y\in S$. Therefore also  $(M,\vec{u}) \sat \langle\vec{A} \gets \vec{a}, \vec{D} \gets \vec{d}, C \gets c' \rangle Y=y\in S$, where $c'=\vec{c'}_C$.

We can now apply the same reasoning we applied to $C$ to some $D \in \vec{D}$, so that by induction we arrive at $(M,\vec{u}) \sat \langle\vec{A} \gets \vec{a}, \vec{C} \gets \vec{c'}  \rangle Y=y \in S$, contradicting our starting point for this case.

$\Leftarrow$: 

We now move on to the implication from right to left. Consider some $(M,\vec{u}) \sat \langle \vec{X} \gets \vec{x}\rangle \phi$. This means there exists some $\vec{v} \sat \phi$ such that 
$(\vec{u},\vec{v})$ is a solution of $M_{\vec{X} \gets \vec{x}}$. So it suffices to show that $(M,\vec{u}) \sat \langle \vec{X} \gets \vec{x}\rangle \V=\vec{v} \in S$, which reduces to showing that $(M,\vec{u}) \sat \langle \vec{X} \gets \vec{x}\rangle \vec{Y}=\vec{y} \in S$, with $\vec{Y} = \V - \vec{X}$ and $\vec{y}=\vec{v}_{\vec{Y}}$.

Per construction of $M$, we have that for each $Y \in \vec{Y}$, $y \in f_Y(\vec{pa_Y})$, where $y=\vec{v}_Y$ and $\vec{pa_Y}=(\vec{u},\vec{v})_{\vec{pa_Y}}$. Let $\vec{A} =\V \cap \vec{Pa_Y}$ and $\vec{B}=\U \cap \vec{Pa_Y}$. As a result, we have that $(M,\vec{u}) \sat \langle\vec{A} \gets \vec{v}_{\vec{A}} \rangle Y=y \in S$. 

For each $Y \in \vec{Y}$, we have that for any $\vec{Z} \subseteq \V - (\vec{A} \cup \{Y\})$: $Z_i \rightarrow Y \not \in \G_S$ for all $Z_i \in \vec{Z}$. Therefore (by the same reasoning as above for $C \in \vec{C}$) for any $\vec{z}$ we have that $(M,\vec{u}) \sat\langle\vec{Z} \gets \vec{z}, \vec{A} \gets \vec{v}_{\vec{A}} \rangle Y=y \in S$. Letting $\vec{Y} = \{Y_1,\ldots,Y_k\}$, we have in particular that for each $i \in \{1,\ldots,k\}$: $(M,\vec{u}) \sat\langle\vec{X} \gets \vec{x}, \vec{Y}^{-i} \gets \vec{y}^{-i} \rangle Y_i=\vec{v}_{Y_i} \in S$, where $\vec{Y}^{-i} := (Y_1, \ldots, Y_{i-1},Y_{i+1},\ldots,Y_k)$ and likewise for $\vec{y}^{-i}$.

Taking $(M,\vec{u}) \sat\langle\vec{X} \gets \vec{x}, \vec{Y}^{-1} \gets \vec{y}^{-1}  \rangle Y_1=y_1 \in S$ and $(M,\vec{u}) \sat\langle\vec{X} \gets \vec{x}, \vec{Y}^{-2} \gets \vec{y}^{-2}  \rangle Y_2=y_2 \in S$, we can apply D5 to derive that  $(M,\vec{u}) \sat\langle\vec{X} \gets \vec{x}, \vec{Y}^{-1,2} \gets \vec{y}^{-1,2} \rangle (Y_1=y_1 \land Y_2=y_2)\in S$. By the same reasoning, we get that $(M,\vec{u}) \sat\langle\vec{X} \gets \vec{x}, \vec{Y}^{-2,3} \gets \vec{y}^{-2,3} \rangle (Y_2=y_2 \land Y_3=y_3)\in S$. Again applying D5 to the last two statements, we get that  $(M,\vec{u}) \sat\langle\vec{X} \gets \vec{x}, \vec{Y}^{-1,2,3} \gets \vec{y}^{-1,2,3} \rangle (Y_1=y_1 \land Y_2=y_2 \land Y_3=y_3)\in S$. By straightforward induction, we get that $(M,\vec{u}) \sat\langle\vec{X} \gets \vec{x} \rangle \vec{Y} \gets \vec{y} \in S$, which is what had to be shown. This concludes the proof.
\eprf

\begin{theorem*}{\bf \ref{thm:red1}} Given models $M_1, M_2, M_3$ over a signature ${\cal S}$ such that $M_2$ is a structural simplification of $M_1$ and $\G_{M_3} = \G_{M_2}$, we have that $M_2$ is an interventional extension of $M_1$, and either $M_3$ is not an interventional extension of $M_1$ or $M_3$ is an interventional extension of $M_2$.\end{theorem*}
  
\prf Assume we have $M_1, M_2, M_3$ as described. First we prove that $M_2$ is an interventional extension of $M_1$. Assume $(M_1,\vec{u}) \sat \langle \vec{X} \gets \vec{x} \rangle \phi$. Just as with the proof of Theorem \ref{thm:exh}, we can assume without loss of generality that $\phi$ is of the form $\V=\vec{v}$. We need to show that $(M_2,\vec{u}) \sat \langle \vec{X} \gets \vec{x} \rangle \V=\vec{v}$. By the theorem from \citep{beckers25} that we stated in the proof of Theorem \ref{thm:part} this reduces to showing that if $(\vec{u},\vec{v})$ is a solution of $M_{1,\vec{X} \gets \vec{x}}$ then $(\vec{u},\vec{v})$ is a solution of $M_{2,\vec{X} \gets \vec{x}}$. 

Consider some $Y \in \V - \vec{X}$, $y \in \R(Y)$, $\vec{a} \in \R(\vec{Pa_Y^1})$, $\vec{b} \in \R(\vec{Pa_Y^2})$, such that $y = \vec{v}_Y$, $\vec{a}=(\vec{u},\vec{v})_{\vec{A}}$, $\vec{b}=(\vec{u},\vec{v})_{\vec{B}}$, and let $g_Y$/$f_Y$ denote the function for $Y$ in $M_1$/$M_2$ respectively. We know that $y \in g_Y(\vec{a})$ and we need to show that $y \in f_Y(\vec{b})$. Per definition of a structural simplification, $f_Y(\vec{b})=g_Y(\vec{b})$. Per definition of a generalized function, $g_Y(\vec{b})=\bigcup\limits_{\{ \vec{a'} | \vec{a'}_{\vec{B}} = \vec{b}  \}} g_Y(\vec{a'}) \supseteq g_Y(\vec{a}) \supseteq \{y\}$. Hence $y \in f_Y(\vec{b})$. This concludes the first part of the proof.

Second we prove that either $M_3$ is not an interventional extension of $M_1$ or $M_3$ is an interventional extension of $M_2$. We do so by a reductio: assume that $M_3$ is an interventional extension of $M_1$ and $M_3$ is not an interventional extension of $M_2$. 

The latter statement means that there is some $(M_2,\vec{u}) \sat \langle \vec{X} \gets \vec{x} \rangle \phi$ and $(M_3,\vec{u}) \sat [\vec{X} \gets \vec{x}]\lnot \phi$. This means there exists some solution $(\vec{u},\vec{v})$ of $M_{2,\vec{X} \gets \vec{x}}$ that is not a solution of $M_{3,\vec{X} \gets \vec{x}}$. 

So there exists some $Y \in \V - \vec{X}$, $y \in \R(Y)$, and $\vec{b} \in \R(\vec{Pa_Y^2})$, such that $y = \vec{v}_Y$, $\vec{b}=(\vec{u},\vec{v})_{\vec{B}}$, $y \in f_Y(\vec{b})$ and $y \not \in h_Y(\vec{b})$. Here $f_Y$ is $Y$'s function in $M_2$, and $h_Y$ is $Y$'s function in $M_3$. 

Letting $g_Y$ denote $Y$'s function in $M_1$, since $M_2$ is a structural simplification of $M_1$, we also have that $f_Y(\vec{b})=g_Y(\vec{b})$ and thus $y \in g_Y(\vec{b})$. Therefore there exists some $\vec{a} \in \R(\vec{Pa_Y^1})$ such that $\vec{a}_{\vec{B}}=\vec{b}$ and $y \in g_Y(\vec{a})$. This implies that for some $\vec{u'} \in \R(\U)$ with $\vec{u'}_{\vec{A}}=\vec{a}_{\U}$ and some $\vec{z} \in \R(\V - \{Y\})$ with $\vec{z}_{\vec{A}}=\vec{a}_{\V}$, we have $(M_1,\vec{u'}) \sat \langle \vec{Z} \gets \vec{z} \rangle Y=y$. 

Per our first assumption, also $(M_3,\vec{u'}) \sat \langle \vec{Z} \gets \vec{z} \rangle Y=y$. Given that $\vec{B} \subseteq \vec{A}$, it follows that $(\vec{u'},\vec{z},y)_{\vec{B}}=\vec{a}_{\vec{B}}=\vec{b}$. Since $(\vec{u'},\vec{z},y)$ is a solution of $M_{3,\vec{Z} \gets \vec{z}}$, this implies that $y \in h_Y(\vec{b})$, contradicting our earlier statement. This concludes the proof.
\eprf

\begin{proposition*}{\bf \ref{pro:ac}} If $X = x$ is an actual cause of $Y=y$ in $(M_1,\vec{u},\vec{v})$ then there exists a structural simplification $(M_2,\vec{u},\vec{v})$ of $(M_1,\vec{u},\vec{v})$ so that $M_2$ is a structural simplification of $M_1$ and $Y=y$ counterfactually depends on $X=x$ in $(M_2,\vec{u},\vec{v})$.
\end{proposition*}

\prf Assume $Y=y$ counterfactually depends on $X=x$ in $(M_2,\vec{u},\vec{v})$, where $M_2^{(\vec{u},\vec{v})}$ is a structural simplification of $M_1^{(\vec{u},\vec{v})}$, but $M_2$ is not a structural simplification of $M_1$. We need to show that there exists some $M_3$ over the same signature ${\cal S}$ such that:
\begin{enumerate}
\item $M_3$ is a structural simplification of $M_1$,
\item  $M_3^{(\vec{u},\vec{v})}$ is a structural simplification of $M_1^{(\vec{u},\vec{v})}$,
\item $(M_3,\vec{u},\vec{v}) \sat (X = x  \land Y=y)$, 
\item $(M_3,\vec{u},\vec{v}) \sat  \langle X \gets x'\rangle Y \neq y$.
\end{enumerate}

Note that $\G_{M_1}=\G_{M_1^{(\vec{u},\vec{v})}}$ and $\G_{M_2}=\G_{M_2^{(\vec{u},\vec{v})}}$, so  $\G_{M_2}$ is a structural simplification of $\G_{M_1}$.

Let $M_3$ be the unique structural simplification of $M_1$ such that $\G_{M_3}=\G_{M_2}$. We claim that $M_3$ also satisfies the remaining three requirements. 

{\bf Condition 3:}

Given that $(M_2,\vec{u},\vec{v})$ is a structural simplification of $(M_1,\vec{u},\vec{v})$, $(\vec{u},\vec{v})$ is a solution of $M_1$. The latter statement is equivalent to $(M_1,\vec{u}) \sat \<\> \V=\vec{v}$. By Theorem \ref{thm:red1}, $M_3$ is an interventional extension of $M_1$, and thus $(\vec{u},\vec{v})$ is also a solution of $M_3$, which implies the result.

{\bf Condition 2:}
 
Consider some $Z \in \V$. We know that $h_Z(\vec{Pa_Z^2}) = g_Z(\vec{Pa_Z^2})$, where $h_Z$/$g_Z$ is $Z$'s function in $M_3$/$M_1$. We know that $f_Z^{(\vec{pa_Z^2},z)}(\vec{Pa_Z^2})=g_Z^{(\vec{pa_Z^1},z)}(\vec{Pa_Z^2})$, where $f_Z$ is $Z$'s function in $M_2$, and $\vec{pa_Z^2}=(\vec{u},\vec{v})_{\vec{Pa_Z^2}}$, $\vec{pa_Z^1}=(\vec{u},\vec{v})_{\vec{Pa_Z^1}}$, $z=\vec{v}_{Z}$. We need to show that $h_Z^{(\vec{pa_Z^2},z)}(\vec{Pa_Z^2}) = g_Z^{(\vec{pa_Z^1},z)}(\vec{Pa_Z^2})$.

First consider some $\vec{pa_Z^2}'\neq \vec{pa_Z^2}$. Then $h_Z^{(\vec{pa_Z^2},z)}(\vec{pa_Z^2}')=h_Z(\vec{pa_Z^2}')$ and $g_Z^{(\vec{pa_Z^1},z)}(\vec{pa_Z^2}')=g_Z(\vec{pa_Z^2}')$, from which the result follows.

Second consider $\vec{pa_Z^2}$. We have that $h_Z^{(\vec{pa_Z^2},z)}(\vec{pa_Z^2})=\{z\}=f_Z^{(\vec{pa_Z^2},z)}(\vec{pa_Z^2})=g_Z^{(\vec{pa_Z^1},z)}(\vec{pa_Z^2})$.

{\bf Condition 4:}

Since both $M_2^{(\vec{u},\vec{v})}$ and $M_3^{(\vec{u},\vec{v})}$ are structural simplifications of $M_1^{(\vec{u},\vec{v})}$, and $\G_{M_2^{(\vec{u},\vec{v})}} =\G_{M_3^{(\vec{u},\vec{v})}}$, it follows that $M_2^{(\vec{u},\vec{v})}=M_3^{(\vec{u},\vec{v})}$, implying the result.\eprf

\end{document}

%% file: defn.tex
\newcommand{\thm}{\begin{theorem}}
\newcommand{\lem}{\begin{lemma}}
\newcommand{\pro}{\begin{proposition}}
\newcommand{\dfn}{\begin{definition}}
\newcommand{\rem}{\begin{remark}}
\newcommand{\xam}{\begin{example}}
\newcommand{\cor}{\begin{corollary}}
\newcommand{\prf}{\noindent{\bf Proof:} }
\newcommand{\ethm}{\end{theorem}}
\newcommand{\elem}{\end{lemma}}
\newcommand{\epro}{\end{proposition}}
\newcommand{\edfn}{\bbox\end{definition}}
\newcommand{\erem}{\bbox\end{remark}}
\newcommand{\exam}{\bbox\end{example}}
\newcommand{\ecor}{\end{corollary}}
\newcommand{\eprf}{\bbox\vspace{0.1in}}
\newcommand{\beqn}{\begin{equation}}
\newcommand{\eeqn}{\end{equation}}
\newcommand{\bbox}{\vrule height7pt width4pt depth1pt}

\newcommand{\clm}{\begin{claim}}
\newcommand{\eclm}{\end{claim}}
\newcommand{\sat}{\models}

\newcommand{\union}{\cup}
\renewcommand{\phi}{\varphi}
\newcommand{\F}{{\cal F}}
\newcommand{\G}{{\cal G}}

\renewcommand{\P}{{\cal P}}

\newcommand{\R}{{\cal R}}

\newcommand{\U}{{\cal U}}
\newcommand{\V}{{\cal V}}

\newcommand{\<}{\langle}
\renewcommand{\>}{\rangle}

\newcommand{\ol}{\setlength{\itemsep}{0pt}\begin{enumerate}}
\newcommand{\eol}{\end{enumerate}\setlength{\itemsep}{-\parsep}}
\newcommand{\ul}{\setlength{\itemsep}{0pt}\begin{itemize}}
\newcommand{\dl}{\setlength{\itemsep}{0pt}\begin{description}}
\newcommand{\edl}{\end{description}\setlength{\itemsep}{-\parsep}}
\newcommand{\eul}{\end{itemize}\setlength{\itemsep}{-\parsep}}